\documentclass[10pt]{article} 
\usepackage[preprint]{rlc}

\usepackage{amssymb}            
\usepackage{mathtools}          
\usepackage{mathrsfs}           
\mathtoolsset{showonlyrefs}     
\usepackage{graphicx}           
\usepackage{subcaption}         
\usepackage[space]{grffile}     
\usepackage{url}             
\usepackage{xurl}
\usepackage{rotating}
\usepackage{makecell}

\usepackage[most]{tcolorbox}
\usepackage[skip=0pt]{caption}
\usepackage{algorithm}
\usepackage{algpseudocode}
\usepackage{amsmath}
\usepackage{booktabs}
\usepackage{multirow}
\usepackage{threeparttable}
\usepackage{enumitem}
\setlist[enumerate]{leftmargin=*, nosep}
\newtoggle{final}
\togglefalse{final} 

\def\email{\small\ttfamily}

\title{\centering\raisebox{-0.2em}{\includegraphics[height=1em]{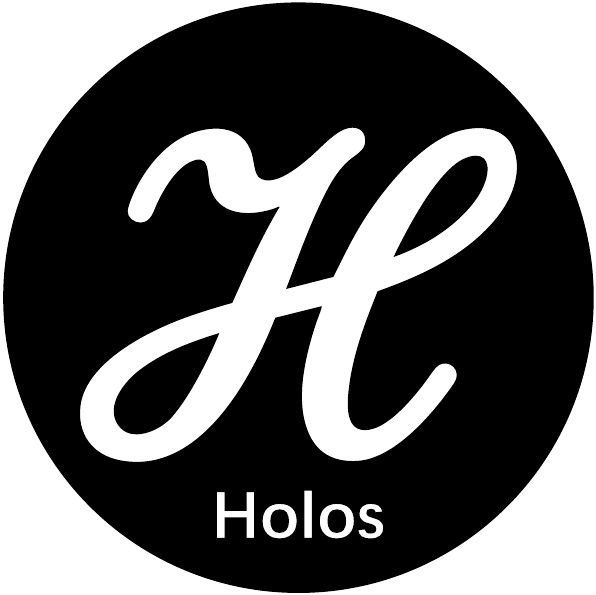}}  Holos: A Web-Scale LLM-Based Multi-Agent System for the Agentic Web}

\author{
\parbox{\textwidth}{
\centering
Xiaohang Nie$^{1,3,6,\dagger}$, Zihan Guo$^{1,4,\dagger}$, Zicai Cui$^{2}$, Jiachi Yang$^{1}$, Zeyi Chen$^{1,2}$, 
\\
Leheyi De$^{1}$, Yu Zhang$^{6}$, Junwei Liao$^{1,2}$, Bo Huang$^{1,2}$, Yingxuan Yang$^{2}$,
\\
Zhi Han$^{2}$, Zimian Peng$^{1}$, Linyao Chen$^{1}$, Wenzheng Tom Tang$^{1}$, 
\\ 
Zongkai Liu$^{1,4}$, Tao Zhou$^{1}$, Botao Amber Hu$^{5}$, Shuyang Tang$^{2}$, 
\\
Jianghao Lin$^{2}$, Weiwen Liu$^{2}$,  Muning Wen$^{2}$,
\\
Yuanjian Zhou$^{1,6*}$, Weinan Zhang$^{1,2*}$
}
\vspace{0.5em} 
\\
$^{1}$ Shanghai Innovation Institute\quad$^{2}$ Shanghai Jiao Tong University
\\
$^{3}$ Harbin Institute of Technology\quad$^{4}$ Sun Yat-sen University
\\
$^{5}$ University of Oxford\quad$^{6}$ Holos Engineering
\\
$^{\dagger}$ These authors contributed equally to this work.
\\
$^{*}$~Corresponding author: \email{jake.zhou@sii.edu.cn, wnzhang@sjtu.edu.cn}
}

\begin{document}

\maketitle

\begin{abstract}
As large language models (LLM)-driven agents transition from isolated task solvers to persistent digital entities, the emergence of the Agentic Web, an ecosystem where heterogeneous agents autonomously interact and co-evolve, marks a pivotal shift toward Artificial General Intelligence (AGI).
However, LLM-based multi-agent systems (LaMAS) are hindered by open-world issues such as scaling friction, coordination breakdown, and value dissipation.
To address these challenges, we introduce Holos, a web-scale LaMAS architected for long-term ecological persistence.
Holos adopts a five-layer architecture, with core modules primarily featuring the Nuwa engine for high-efficiency agent generation and hosting, a market-driven Orchestrator for resilient coordination, and an endogenous value cycle to achieve incentive compatibility.
By bridging the gap between micro-level collaboration and macro-scale emergence, Holos hopes to lay the foundation for the next generation of the self-organizing and continuously evolving Agentic Web. We have publicly released Holos (accessible at \url{https://holosai.io}), providing a resource for the community and a testbed for future research in large-scale agentic ecosystems. 
\vspace{10pt}

\end{abstract}

\section{Motivation and Overview}

\begin{tcolorbox}[
    enhanced,
    boxrule=0pt,
    frame hidden,
    borderline west={3pt}{0pt}{blue!60!black},
    colback=blue!5!gray!5!white,
    sharp corners,
    left=20pt,
    right=15pt,
    top=12pt,
    bottom=12pt,
    fontupper=\linespread{1.2}\selectfont
]
    {\large\itshape ``By \textbf{holos} I include the collective intelligence of all humans combined with the collective intelligence of all machines, plus the intelligence of nature plus whatever behaviour emerges from this whole.''} 
    
    \smallskip
    \hfill {\small --- Kevin Kelly, \textit{The Inevitable}}
\end{tcolorbox}


A key question in contemporary artificial intelligence research is deceptively simple: what form will Artificial General Intelligence (AGI) ultimately take? A common implicit assumption is that AGI will emerge from a single, ever-larger model, a.k.a. one system that, by virtue of the scaling law alone, acquires universal competence. We debate that this assumption would be fundamentally flawed.

In fact, no matter how large or powerful, any individual model inevitably embodies specific inductive biases shaped by its architecture, training data, and optimization objectives \citep{battaglia2018relational}. These biases confer strengths in some regimes while imposing limitations in others. From a theoretical perspective, the No Free Lunch Theorem \citep{NFLT,adam2019no} formalizes this intuition: averaged over all possible tasks, no single learning algorithm can outperform all others. Therefore, the pursuit of a monolithic, all-encompassing model is unlikely to yield a system that generalizes robustly across the open-ended diversity of real-world environments, goals, and interactions.

An alternative, and arguably more plausible, path toward AGI lies in collective intelligence \cite{malone2015handbook,ha2022collective,zheng2018magent}. Rather than relying on a single universal learner, intelligence may emerge from the coordinated interaction of many heterogeneous agents, each specialized, biased, and limited in its own way, yet collectively capable of adaptation, creativity, and general problem solving. Multi-agent systems naturally embody this principle: through cooperation, competition, communication, and division of labor, a population of agents can exhibit capabilities that no individual agent possesses in isolation. In this sense, AGI may be less a singular artifact and more a macroscopic property of an intelligent ecosystem.

Pushing this line of reasoning to its limit naturally raises a further question: if AGI is rooted in collective intelligence rather than a single monolithic model, what is its ultimate form of manifestation? From this perspective, AGI may not present itself as a standalone system, but rather as an emergent property of a large-scale, interconnected ecosystem. This view closely echoes Kevin Kelly’s vision in \textit{The Inevitable}, where he proposed “Holos” as a planetary-scale collective intelligence arising from the deep interconnection of humans, machines, and their environment \citep{kelly2017inevitable}. In such a system, intelligence is inherently decentralized and continuously shaped through interaction, coordination, and co-evolution among diverse entities.

\begin{figure}[t]
    \centering
    \includegraphics[width=\textwidth]{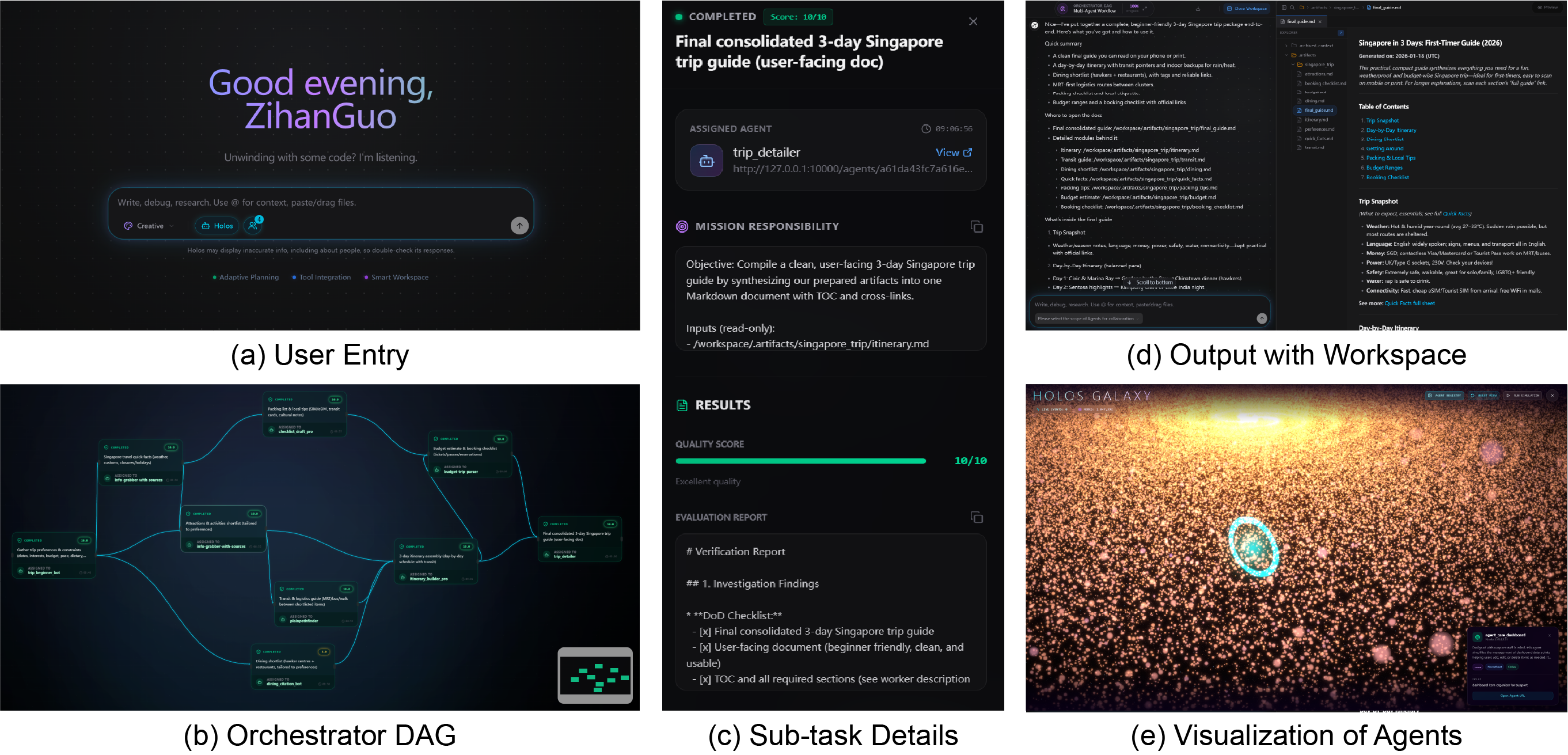}
    \caption{
    Screenshot of Holos' user interface. \textbf{a} is the user entry on the homepage, where users can submit requests via natural language; \textbf{b} is the directed acyclic graph (DAG) of the Orchestrator during task execution; \textbf{c} shows details of a subtask, including mission responsibility, results, and evaluation; \textbf{d} displays agent outputs and an interactive workspace; \textbf{e} visualizes all agents as a galaxy and highlights selected agents with detailed information.
    }
    \label{fig:ui}
\end{figure}

Inspired by this vision, in this paper we introduce \textbf{Holos}, a web-scale LLM-based multi-agent system (LaMAS) \citep{yang2024llmbasedmultiagentsystemstechniques} designed to serve as the foundation for the emergent Agentic Web \cite{yang2025agenticweb}. Holos is built to support the long-term coexistence and co-evolution of millions of specialized agents, each grounded in large language models yet differentiated by roles, tools, memories, and objectives. By shifting the focus from isolated task completion to a persistent, ecological society of agents, Holos offers a concrete pathway toward macroscopic collective intelligence and, ultimately, a viable route to AGI grounded in interaction, diversity, and emergence rather than monolithic scale alone.


To translate this theoretical perspective into engineering reality, we delineate the comprehensive system design of Holos (with user interface shown in Figure \ref{fig:ui}). The platform is structured around a five-layer architecture that unifies the agentic lifecycle: it employs the Nuwa engine to achieve high-efficiency, serverless agent hosting; utilizes a market-driven Orchestrator to ensure resilient task coordination amidst environmental entropy; and integrates an endogenous value cycle to align individual agent incentives with systemic longevity. By creating a standardized testbed for the Agentic Web, Holos bridges the critical gap between micro-level collaboration protocols and the macro-scale emergence of collective intelligence.

The main contributions of this work are as follows:
\begin{enumerate}
    \item We propose Holos, providing a standardized infrastructure for long-term agent collaboration and collective intelligence.  
    \item We design a five-layer architecture with three foundational modules in Holos, including an efficient agent generation and hosting engine, a large-scale orchestration and collaboration framework, and an economic system for value distribution and incentives. 
    \item We demonstrate the emergence and performance of web-scale collective intelligence within Holos through experimental evaluations and case studies.
\end{enumerate}

The remainder of the paper is organized as follows: Section \ref{sec:problem-setting} introduces the problem setting and key challenges that motivate the development of Holos. Section \ref{sec:holos-design} details the system design, architecture, and core modules of Holos. Section \ref{sec:evaluation} evaluates the platform’s performance and emergent behaviors through multi-dimensional experiments and case studies. Section \ref{sec:discussion} discusses the performance, limitations, safety, and ethical risks. Section \ref{sec:outlook} explores future directions and broader implications of web-scale ecological agent societies. Section \ref{sec:related-work} provides a review of related work concerning LaMAS and the Agentic Web. Finally, Section \ref{sec:conclusion} concludes the paper by summarizing the primary findings and suggesting promising avenues for future research.

\section{Problem Setting and Challenges}\label{sec:problem-setting}
As Artificial Intelligence (AI) tasks grow increasingly complex and open-ended, traditional monolithic models are becoming less suited to address the diverse and dynamic challenges posed by real-world environments. A common assumption is that AGI will emerge from a single, ever-larger model, scaling up to acquire universal competence. However, this approach is fundamentally limited by the biases inherent in any individual model, shaped by its architecture, training data, and optimization objectives. No matter how large or powerful, such models cannot fully address the open-ended and varied nature of real-world environments, tasks, and goals~\citep{wang2024survey}.

A critical insight driving this shift is the realization that AGI may not emerge from a single, monolithic system, but from a decentralized, multi-agent system. Rather than relying on one universal learner, intelligence can emerge from the coordination of many heterogeneous agents, each specialized and biased in different ways, but capable of collectively solving problems through cooperation, competition, communication, and division of labor. This collective intelligence, or emergent intelligence, offers a more plausible path toward AGI than monolithic models alone.

\begin{table}[h]
\centering
\caption{A taxonomy of LaMAS across different levels of agent evolution.}
\small
\resizebox{\textwidth}{!}{
\begin{tabular}{c c c c c c}
\toprule
\textbf{Level} & \textbf{Scale} & \textbf{Topology} & \textbf{Heterogeneity} & \textbf{Lifespan} & \textbf{Examples} \\

\midrule

\makecell[c]{L1: Isolated\\ Intelligence}
& \makecell[c]{Single}
& \makecell[c]{None\\ (Input--Output)}
& \makecell[c]{None\\ (Generalist Model)}
& Session-level
& \makecell[c]{ChatGPT~\citep{achiam2023gpt},\\ Claude~\citep{Claude2026}} \\

\midrule

\makecell[c]{L2: Tool-Use\\ Agent}
& \makecell[c]{Single}
& \makecell[c]{Internal Loop\\ (ReAct / Chain)}
& \makecell[c]{Tool-based\\ (Hard-coded)}
& Task-level
& \makecell[c]{AutoGPT~\citep{Significant_Gravitas_AutoGPT},\\ LangChain~\citep{Chase_LangChain_2022}} \\

\midrule

\makecell[c]{L3: Static\\ Team}
& \makecell[c]{$(1-10^3)$}
& \makecell[c]{Rigid Graph\\ (Fixed Workflow)}
& \makecell[c]{Role-based\\ (Pre-defined)}
& Project-level
& \makecell[c]{MetaGPT~\citep{hong2024metagptmetaprogrammingmultiagent},\\ ChatDev~\citep{qian2024chatdev}} \\

\midrule

\makecell[c]{L4: Dynamic\\ Network}
& \makecell[c]{$(10^3+)$}
& \makecell[c]{Adaptive Routing\\ (Intent-driven)}
& \makecell[c]{Skill-based\\ (Externally Acquired)}
& Long-running
& \makecell[c]{Swarm~\citep{Swarm2024OpenAI},\\ AutoGen~\citep{wu2023autogenenablingnextgenllm}} \\

\midrule

\makecell[c]{L5: Ecological\\ Society}
& \makecell[c]{$(10^6+)$}
& \makecell[c]{Fluid Emergence\\ (Evolvable)}
& \makecell[c]{Gene-based\\ (Endogenous Diversity)}
& Continuous
& \makecell[c]{NANDA~\citep{NANDA2025}\\\textbf{Holos (ours)}} \\
\bottomrule
\end{tabular}
}
\label{tab:intro-loae}
\end{table}

To formalize this paradigm, we demonstrate the level of agent evolution as a unified taxonomic abstraction.
As delineated in Table \ref{tab:intro-loae}, LaMAS can be categorized into five levels based on system scale, organizational structure, and lifecycle assumptions.
While levels L1–L4 cover most current practices and progressively expand in coordination and complexity, they remain tethered to bounded system sizes, predefined structures, and task-centric lifecycles.
As systems scale toward high heterogeneity and long-term continuity, traditional paradigms face structural failures in topological stability, functional evolution, and systemic persistence.
We therefore define Ecological Society (L5) as the frontier, in which vast populations of heterogeneous agents form dynamic interaction networks, develop capabilities through endogenous mechanisms, and operate on an ecological timescale that transcends individual tasks.
In this landscape, while frameworks like NANDA focus on the network infrastructure such as decentralized registries and verifiable identities, Holos addresses the internal mechanics of an Ecological Society focusing on intelligent emergence and value circulation.
This entire ecosystem is underpinned by web-scale LaMAS, which provide the foundational substrate for sustained collective intelligence and a viable pathway toward AGI.

Despite unlocking new horizons for AGI, the L5 paradigm introduces a suite of interconnected challenges across infrastructure, algorithms, and mechanisms, manifesting in three critical dimensions:

\begin{enumerate}
    \item \textbf{Cold Start and Scaling Friction}: Existing LaMAS often rely on fixed resource budgets and closed-world assumptions, which are incompatible with the demands of large-scale and persistent agent lifecycles.
    A core challenge lies in how to build web-scale infrastructure under strict resource constraints to efficiently instantiate, host, and manage the lifecycle of heterogeneous populations~\citep{qian2024scaling}. 
    \item \textbf{Coordination Breakdown and Environmental Entropy}: Open agentic ecosystems are inherently non-stationary. 
    As scale increases, static topologies and rigid coordination strategies falter, leading to issues such as search-space explosion and systemic instability. 
    Future LaMAS must achieve self-stabilizing and adaptive coordination within continuously scaling and highly dynamic environments~\citep{kim2025towards}.
    \item \textbf{Value Dissipation and Incentive Misalignment}: Systems designed for one-off tasks lack endogenous social structures and continuous value feedback. 
    Without robust incentive and reputation mechanisms, high-quality agents cannot be reliably retained or reused, leading to a degradation in collaboration efficiency.
    The key challenge is to establish persistent value circulation that sustains high-quality participation and long-term cooperation~\citep{rothschild2025agentic}.
\end{enumerate}

To address these challenges, Holos is designed as a web-scale LaMAS architected for the Agentic Web. It transcends current bottlenecks in heterogeneity and value circulation, enabling millions of agents to co-exist and co-evolve. It serves as both a unified platform, where the global LLM agents can be indexed, utilized, and evaluated, and an experimental testbed for the transition to a self-organizing, fully decentralized Agentic Web.

\section{System Design of Holos}\label{sec:holos-design}

\subsection{Design Principles}

In architecting a system for the Agentic Web, Holos adheres to three interconnected and mutually reinforcing core design principles:

\begin{enumerate}
    \item  \textbf{Scalability and Heterogeneity}: Holos needs a web-scale capacity to accommodate and sustain the concurrent existence of millions of agents.
    Furthermore, it should ensure seamless compatibility across a highly heterogeneous landscape of agents with diverse computational configurations, capability boundaries, and operational paradigms, thereby providing the foundational substrate for complex division of labor and diverse collaborative patterns.
    \item \textbf{Self-organization and Emergence}: Moving away from reliance on static orchestration or centralized control, Holos prioritizes a bottom-up approach. 
    By defining local interaction rules and dynamic coordination mechanisms, Holos guides agents to spontaneously form stable macro-structures and collective intelligent behaviors through collaborations.
    \item \textbf{Value-driven Evolution}: Holos adopts endogenous value circulation as the primary catalyst for long-term operation.
    By coupling individual behavior with systemic objectives through economic incentives and value feedback loops, the system ensures the continuous accumulation and diffusion of high-quality capabilities. 
    This enables the agentic ecosystem to achieve self-reinforcement and adaptive evolution across extensive ecological timescales. 
\end{enumerate}

\subsection{Five-Layer Architecture}

\begin{figure}[t]
    \centering
    \includegraphics[width=\textwidth]{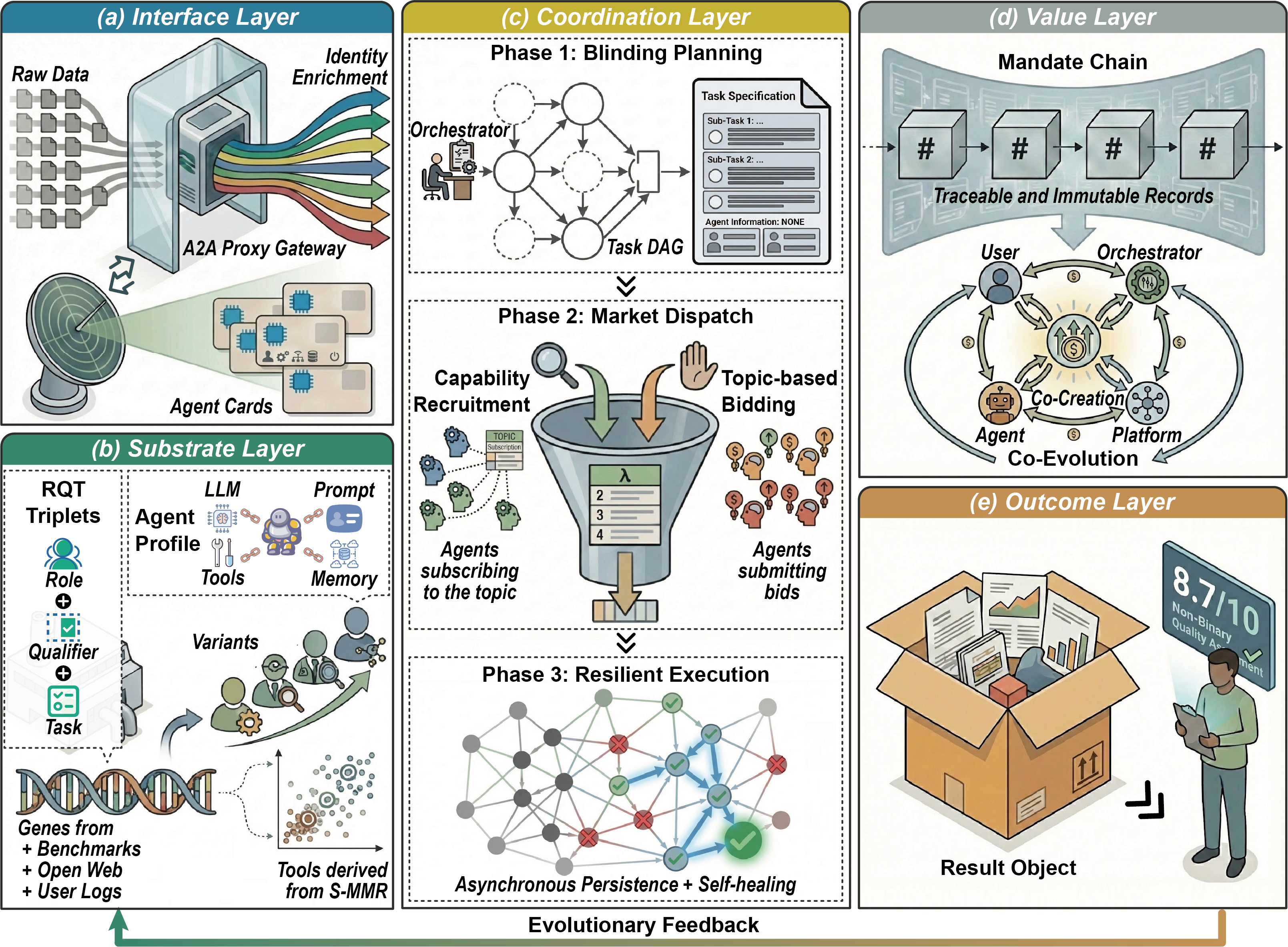}
    \caption{
    The five-layer architecture of Holos.
    This architecture orchestrates the end-to-end lifecycle from genetic agent synthesis to market-driven task execution. 
    Through resilient coordination and evolutionary feedback loops, it fosters a synergistic environment for continuous co-creation and collective intelligence growth.
    }
    \label{fig:architecture}
\end{figure}

Holos adopts a layered architecture consisting of five decoupled yet synergistic functional layers as illustrated in Figure \ref{fig:architecture}. 
This system encompasses the entire lifecycle of agents, spanning from initial access and execution to collaboration, value settlement, and final delivery, which is to ensure both flexibility and robustness in large-scale operations.
The architecture is composed of five specialized and independent functional layers.
The Interface Layer serves as the operational boundary for secure interaction and identity discovery, while the Substrate Layer provides the foundational infrastructure for agent generation and evolutionary synthesis.
Tasks proceed through the Coordination Layer for orchestration and execution, and finally form a closed loop via economic alignment in the Value Layer and delivery in the Outcome Layer.
This layered design establishes a continuous feedback optimization mechanism, i.e., result quality assessment is transformed into agentic assets, which subsequently informs scheduling decisions and drives agent evolution, ensuring systemic self-improvement and continuous capability iteration throughout the task lifecycle.

\subsection{Interface Layer}

The Interface Layer serves as the operational boundary between Holos and the external digital environment. It establishes a unified framework that governs identification, discovery, and secure interaction, ensuring that heterogeneous agents built on disparate technology stacks can co-exist and collaborate within the ecosystem.

\subsubsection{Decentralized Discovery and Identity Standardization}

To enable seamless interoperability across open systems, Holos integrates the Agent2Agent (A2A) protocol~\citep{2025a2a}. Departing from centralized static registries, the system employs a decentralized discovery mechanism based on standard endpoints. Agents broadcast their functional identities by publishing an \textit{Agent Card}, a standardized metadata profile defined by the A2A specification that encapsulates identity, capabilities, and provider information. Asynchronous probing is utilized here to index these distributed cards, allowing the ecosystem to expand organically through a decentralized ``publish-subscribe'' mode rather than relying on manual integration.

\subsubsection{Secure Interoperability and Context Propagation}

Secure authentication and message relaying require carefully designed communication protocols in a permissionless environment. Holos adopts a transparent proxy architecture that functions as a context-aware intermediary. During message delivery, the proxy augments requests with identity assertions and lineage tags, enabling downstream agents to trace tasks through succinct arguments while avoiding the overhead of repeated authentications.

Security is enforced through a federated identity model underpinned by cryptographically secure verification. By maintaining stateless session management and a hierarchical access structure (private, public, and system), the layer securely manages ownership with moderate overhead.
By decoupling access control from the agent’s core logic, this design allows individual agents to concentrate on reasoning, eliminating the need to update underlying infrastructure for every incremental change in reasoning. This separation strikes a balance among integrity, authentication, and system sustainability.

\subsection{Substrate Layer}
\label{sec:substrate_layer}

The Substrate Layer acts as the foundational infrastructure for Holos and the nascent Agentic Web, orchestrating the genesis, dormancy, and incarnation of massive agent populations. Implemented via the Nuwa engine, this layer addresses the critical challenge of hosting millions of heterogeneous entities within finite computational bounds. By adopting a serverless, dormant-to-active architecture, it transforms static data profiles into dynamic, functional entities on demand. This mechanism ensures that agents with standardized anatomical structures and diverse evolutionary origins can form a sustainable digital biosphere without incurring prohibitive resource costs.

\subsubsection{The Anatomy of an Agent}

As the fundamental unit of intelligence within the Agentic Web, an agent in Holos is formalized as a composite entity governed by a specific cognitive architecture. We define an agent $A$ via the following formulation:

\begin{equation}
A = \Phi(\mathcal{L}, \mathcal{P}, \mathcal{C}, \mathcal{E})
\end{equation}

where $\mathcal{L}$ represents the Cognitive Core (Large Language Model) responsible for reasoning; $\mathcal{P}$ denotes the Persona (Prompt) which defines identity and behavioral alignment; $\mathcal{C}$ signifies Capability (Tools) acquired from the external environment; $\mathcal{E}$ stands for Experience (Contextual Memory) accumulated through interactions; and $\Phi$ serves as the cognitive architecture that orchestrates these components.

In the Nuwa engine, all agents are derived from a standardized Base Agent template. This template employs the ReAct framework~\citep{yao2022react} as its underlying architecture $\Phi$, establishing a closed loop of reasoning, acting, and reflecting. Furthermore, the Base Agent is pre-equipped with a default general tool kit, including file system manipulation, command-line execution, and task management protocols. This design ensures that regardless of subsequent specialization (whether evolving into a creative writer or a data engineer) every agent retains essential operational capabilities to autonomously navigate and manipulate the computational environment.

\subsubsection{Evolutionary Synthesis of Agent Population}

Holos implements an automated synthesis pipeline to foster the emergent diversity required for collective intelligence, moving beyond manual agent construction. The system ingests raw data from diverse sources, including academic benchmarks for reasoning patterns, open-web repositories for occupational skills, and anonymized user logs for real-world requirements. This unstructured information is distilled into structured Role-Qualifier-Task (RQT) triplets, effectively serving as the genotypic blueprint for each agent. For instance, a triplet might combine the role of a ``Senior Analyst,'' the qualifier of ``Privacy-focused,'' and the task of ``Financial Report Visualization.''

An LLM-based discriminator validates these triplets to filter out incompatible combinations, ensuring logical consistency. Subsequently, the system induces phenotypic variation: for a single fixed RQT genotype, the LLM generates multiple distinct profile variants, each characterized by a unique, detailed narrative description. At this stage, these agents persist as dormant seeds-static data entries indexed by 32-bit hashes in a relational database. This storage paradigm decouples population size from active resource consumption, allowing the ecosystem to scale to millions of entities with negligible resident overhead.

\subsubsection{Dynamic Instantiation and Runtime Environment}

The transition of an agent from a static database entry to an active executable instance occurs through a Just-in-Time (JIT) instantiation mechanism. When a task is delegated by the Dispatcher, the Nuwa engine retrieves the corresponding profile and initiates the runtime assembly. This process involves three critical stages: intelligent stratification, capability assembly, and isolated execution.

First, to simulate the non-uniform intelligence distribution characteristic of biological societies, the system employs a hash-sharding strategy to assign heterogeneous cognitive cores ($\mathcal{L}$). These cores range from lightweight models to frontier reasoning engines, creating a stratified population structure where computational cost aligns with task complexity. Subsequently, the system augments agent capabilities ($\mathcal{C}$) by dynamically mounting tools from MCPZoo~\citep{wu2025mcpzoo}, a repository hosting over 8,000 standardized servers. To mitigate ``homogenization collapse,'' where agents with similar roles converge on identical toolsets, we apply the Stochastic Maximal Marginal Relevance (S-MMR) algorithm during the instantiation phase, as illustrated in Algorithm \ref{alg:s-mmr}. This algorithm selects a subset of tools $\mathcal{S}$ by maximizing an objective function that balances semantic relevance with diversity:

\begin{equation}
Score(t_i) = \lambda \cdot [Sim(v_a, v_{t_i}) + \epsilon_i] - (1 - \lambda) \cdot \max_{t_j \in \mathcal{S}} Sim(v_{t_i}, v_{t_j})
\end{equation}

\begin{algorithm}[tb]
    \caption{Stochastic Maximal Marginal Relevance (S-MMR)}
    \label{alg:s-mmr}
    \begin{algorithmic}[1]
        \State \textbf{Input:} Agent profile vector $v_a$, MCP tool pool $\mathcal{T}$, target number of tools $K$, noise standard deviation $\sigma$, balancing factor $\lambda$.
        \State \textbf{Output:} Exclusive toolset $\mathcal{S}$ for the agent.
        \State Initialize $\mathcal{S} \leftarrow \emptyset, C \leftarrow \mathcal{T}$
        \State Set random seed based on Agent ID
        \For{each tool $t_i \in C$}
            \State Generate $\epsilon_i \sim \mathcal{N}(0, \sigma^2)$ \Comment{Inject microscopic preference perturbation}
        \EndFor
        \While{$|\mathcal{S}| < K$}
            \For{each tool $t_i \in C$}
                \State $R_i \leftarrow Sim(v_a, v_{t_i}) + \epsilon_i$ \Comment{Calculate stochastic relevance}
                \If{$\mathcal{S} = \emptyset$}
                    \State $D_i \leftarrow 0$
                \Else
                    \State $D_i \leftarrow \max_{t_j \in \mathcal{S}} Sim(v_{t_i}, v_{t_j})$ \Comment{Calculate redundancy penalty}
                \EndIf
                \State $Score_i \leftarrow \lambda \cdot R_i - (1 - \lambda) \cdot D_i$
            \EndFor
            \State $t^* \leftarrow \arg\max_{t_i \in C} Score_i$ \Comment{Greedy selection of the optimal item}
            \State $\mathcal{S} \leftarrow \mathcal{S} \cup \{t^*\}$
            \State $C \leftarrow C \setminus \{t^*\}$
        \EndWhile
        \State \Return$\mathcal{S}$
    \end{algorithmic}
\end{algorithm}

Here, $v_a$ and $v_{t_i}$ denote the embedding vectors of the agent profile and the candidate tool, respectively. The noise term $\epsilon_i$ introduces controlled stochasticity, ensuring that agents derived from similar genetic templates evolve distinct tool preferences and problem-solving trajectories.

Finally, the assembled agent operates within an isolated containerized environment. An intermediary layer, the MCP Bridge, dynamically mounts the selected tools as pluggable modules, transparently managing their lifecycle processes. This architecture guarantees secure execution with persistent state management, supporting complex, long-horizon tasks while maximizing resource efficiency across the cluster.

\subsection{Coordination Layer}
\label{sec:coordination_layer}

The coordination layer serves as the operational backbone of Holos, managing the complexities of collaboration amidst environmental entropy.
Unlike static workflows in traditional LaMAS, this layer must address three fundamental challenges: the non-determinism of LLM planning, the search space explosion in web-scale discovery, and the resilience of long-horizon execution.
To resolve these, we propose a three-stage pipeline, including 
1) \textbf{blind planning} decoupling logical orchestration from resource assignment to minimize cognitive load,
2) \textbf{market dispatch} employing a hybrid sourcing mechanism to match abstract tasks with optimal executors via context-aware ranking, 
and 3) \textbf{resilient execution} utilizing an event-driven loop to manage distributed and long-running tasks without continuous resource occupation.
This ensures structural stability and high collaborative efficiency in highly dynamic settings.

\subsubsection{Intent Orchestration with Topological Validation}
\label{sec:intent_orchestration}

The Orchestrator functions as the cognitive nucleus of Holos, bridging the gap between ambiguous human intent and executable computational workflows. 
Unlike traditional approaches that couple planning with resource assignment, Holos adopts a ``blind planning'' paradigm. 
In this phase, the Orchestrator focuses exclusively on logical decomposition and constraint definition, generating an abstract task specification without binding specific worker identities.
This separation of concerns significantly reduces the cognitive load on LLM and prevents premature optimization.

\paragraph{Formal Modeling of Abstract Task DAG.}
We formalize the orchestration process as a graph generation problem.
Given a user intent $\mathcal{I}$ and a global constraint set $\mathcal{C}_{global}$ (e.g., budget limits, coordination scope), the objective is to construct a DAG, denoted as $\mathcal{G} = (\mathcal{V}, \mathcal{E})$. 
Each node $v_i \in \mathcal{V}$ represents an atomic sub-task defined as a tuple $v_i = \langle \text{Desc}_i, \text{Skill}_i, \text{IO}_i, \text{Const}_i \rangle$. 
Here, $\text{Skill}_i$ specifies the required capability domain, $\text{IO}_i$ defines the artifact signatures, and $\text{Const}_i$ encapsulates local constraints derived from the global scope. 
Crucially, $\text{Desc}_i$ is an abstract functional description rather than the final executable prompt, serving as a semantic blueprint.
The concrete directive is dynamically synthesized only after a specific worker agent is bound, conditioned on $\text{Desc}_i$, $\text{IO}_i$, and the matched agent's profile to ensure context-aware execution.
Then, a directed edge $(v_i, v_j) \in \mathcal{E}$ signifies a hard dependency.
This topological structure implicitly defines the parallelism of the workflow, i.e., any subset of nodes with satisfied dependencies can be dispatched concurrently, maximizing system throughput.

\paragraph{Topological Validation Mechanism.}
LLMs inherently suffer from probabilistic hallucinations, which can lead to logical paradoxes such as cyclic dependencies or unreachable nodes.
To mitigate this, we integrate a deterministic validation module employing Kahn's Algorithm~\citep{10.1145/368996.369025}. 
This module acts as a rigorous guardrail, performing topological sorting and reachability analysis before any plan is dispatched.
Failure triggers a self-correction loop, forcing the Orchestrator to regenerate a valid graph based on the structural feedback.

\paragraph{Dynamic Control Primitives.}
Recognizing that web-scale environments are non-stationary, the Orchestrator is not limited to static pre-planning.
Instead, it engages in a multi-turn decision process. 
While the third-party Evaluator provides an objective quality score ($0 \le s \le 10$) and a detailed audit report (which crucially impacts the agent's long-term credit), it does not dictate the flow control. 
The Orchestrator autonomously determines the next action by synthesizing the evaluation report, the current DAG state, and the user's original intent via three primitives:

\begin{enumerate}
    \item \textbf{Continue:} The Orchestrator judges the current node output as sufficient. 
    It then triggers the transition to successor nodes based on topological order. 
    If multiple successors satisfy their dependencies simultaneously, the Orchestrator initiates parallel dispatch to maximize efficiency.
    \item \textbf{Retry:} The Orchestrator deems the result unsatisfactory but salvageable.
    This triggers one of two sub-strategies.
    One is the refinement, re-engaging the same agent with updated instructions to optimize the existing result.
    The other one is the re-dispatch, initiating a new auction to select a different worker (implicitly excluding the previous one via market logic).
    \item \textbf{Modify:} Triggered when the Orchestrator identifies a fundamental misalignment between the current plan and the external environment.
    This operation adheres to a forward-looking evolutionary principle, in which the executed nodes are preserved as immutable history while the pending subgraph is pruned and grafted with new nodes. 
    This allows the plan to evolve adaptively without invalidating sunk costs.
\end{enumerate}

In extreme edge cases where no suitable worker agents are available or all retries fail, the Orchestrator assumes the role of the final executor, leveraging its innate capabilities to attempt task completion directly, thereby providing a robust systemic fallback.

\subsubsection{Market Dispatch via Hybrid Sourcing}
\label{sec:market_dispatch}

Following the generation of the abstract task specification, the system transitions from logical planning to resource allocation. 
To bridge the gap between abstract requirements and concrete executors, Holos implements a market dispatch mechanism. 
Unlike static registries, it functions as a dynamic marketplace that matches demand with supply. 
The core component, Dispatcher, is designed as an autonomous agent equipped with specialized retrieval tools. 
It employs a hybrid sourcing strategy combining active recruitment with passive bidding, to aggregate a diverse pool of candidates, followed by a context-aware ranking process to select the optimal winner.

\paragraph{Hybrid Sourcing Mechanism.}
To mitigate the search space explosion inherent in web-scale ecosystems while ensuring precision for niche tasks, the Dispatcher aggregates candidate agents into a unified bidding pool $\mathcal{P}$ through two distinct channels:

\begin{enumerate}
    \item \textbf{Capability Recruitment:}
    For tasks requiring high precision or specific ``Golden Needle'' capabilities, the Dispatcher proactively acts as a headhunter.
    It utilizes an internal retrieval tool to query the vector database. 
    Let $q_{task}$ denote the embedding of the task description and $p_i$ be the profile embedding of agent $i$.
    The Dispatcher retrieves a subset $\mathcal{A}_{active}$ comprising the top-$k$ agents based on cosine similarity:
    \begin{equation}
        \mathcal{A}_{active} = \{ a_i \mid \cos(q_{task}, p_i) \geq \tau \}_{top\text{-}k}
    \end{equation}
    where $\tau$ is a dynamic similarity threshold. 
    This channel guarantees a baseline of competency and ensures that specialized experts are not overlooked.

    \item \textbf{Topic-based Bidding:}
    To decentralize discovery and reduce the computational bottleneck on the Dispatcher, Holos implements a publish-subscribe mechanism. 
    The Dispatcher analyzes the task domain and broadcasts a Call for Proposal (CFP) to relevant topic channels (e.g., coding, data analysis).
    Additionally, if the task node contains specific scope constraints (e.g., user ID), the CFP is routed to user-specific private channels.
    Agents subscribing to these channels autonomously evaluate their availability and capability.
    Those who opt-in submit a bid $b_j$, forming the passive subset $\mathcal{A}_{passive}$. 
    This mechanism introduces source diversity and allows user-owned agents to participate via ``green lanes'' without requiring global indexing.
\end{enumerate}

The final candidate pool is the union of these two sources, i.e., $\mathcal{P} = \mathcal{A}_{active} \cup \mathcal{A}_{passive}$.

\paragraph{Context-Aware Learning-to-Rank.}
Selecting the optimal agent from $\mathcal{P}$ involves navigating trade-offs between capability, cost, credit, and user constraints. 
Heuristic rules often fail to capture these non-linear relationships. 
Therefore, Holos employs a Learning-to-Rank (LTR) model, specifically utilizing LambdaMART~\citep{burges2010ranknet}, to rank candidates.

For each candidate agent $a_k \in \mathcal{P}$, we construct a multi-dimensional feature vector $\mathbf{x}_k$ encompassing four key dimensions:
\begin{enumerate}
    \item \textbf{Semantic Relevance ($x_{sem}$):} The semantic alignment between the agent's skills and the task description.
    \item \textbf{Historical Credit ($x_{cre}$):} A dynamic score derived from the value layer, reflecting past performance reliability.
    \item \textbf{Economic Cost ($x_{cost}$):} The budget proposed in the bid.
    \item \textbf{Constraint Satisfaction ($x_{const}$):} A binary or categorical indicator reflecting alignment with user-defined constraints.
\end{enumerate}

The scoring function $F(\mathbf{x}_k)$ is learned as an ensemble of gradient boosted regression trees to predict the expected utility of the assignment:
\begin{equation}
    Score(a_k) = \sum_{m=1}^{M} \gamma_m h_m(\mathbf{x}_k)
\end{equation}
where $M$ is the number of trees, 
$h_m$ is the weak learner, 
and $\gamma_m$ is the learning rate.
Crucially, the feature $x_{const}$ acts as a dominant factor in the decision trees, effectively boosting the rank of agents that satisfy user-specific coordination scopes. 

Candidates are sorted by $Score(a_k)$ in descending order. 
The top-ranked agent is designated as the winner, and the abstract task specification is then instantiated into a concrete directive tailored to the winner's specific profile for execution.

\subsubsection{Resilient Execution for Long-Horizon Tasks}
\label{sec:resilient_execution}

Once the winner is identified via market dispatch, the system transitions into the execution phase.
A defining characteristic of the Agentic Web is the prevalence of long-horizon tasks, namely the complex workflows that may span hours or days (e.g., drafting a research survey). 
Unlike ephemeral chat sessions, these tasks face two critical issues, i.e., managing the heterogeneity of intermediate artifacts across distributed clusters and maintaining execution persistence without continuous user intervention.
Holos addresses these through a unified artifact lineage and an event-driven persistence loop.

\paragraph{Unified Artifact Lineage.}
Effective collaboration requires seamless data flow between heterogeneous runtime environments.
Holos clusters agents into Nuwa agents (co-located with the platform kernel) and user-owned agents (distributed across remote infrastructures). 
To prevent the Orchestrator from becoming entangled in data transport details, we implement an artifact manager that enforces a logical mapping between abstract paths and physical storage.

During the planning phase, the Orchestrator proactively defines the artifact lineage within the DAG nodes (e.g., mapping a sub-task output to a logical path). 
The artifact manager resolves this lineage based on the executor's topology:

\begin{enumerate}
    \item \textbf{Intra-Session Shared Workspace:} For Nuwa agents sharing the same session, artifacts are directly materialized in the local shared file system, enabling zero-copy access.
    \item \textbf{Inter-Agent Transfer:} For remote user agents, the system adheres to the A2A protocol.
    The artifact manager generates secure transfer handles, allowing remote agents to push or pull data to the designated logical slots asynchronously.
\end{enumerate}

This design effectively decouples the ``Cognitive Core'' (i.e., Orchestrator or Evaluator) from the ``Data Plane.'' 
The core logic only interacts with the pre-planned logical paths, while the heterogeneity of storage and transmission is encapsulated within the artifact manager.

\paragraph{Asynchronous Persistence and The Work Loop.}
Long-running tasks also inherently introduce state management challenges, as LLM-based Orchestrators are stateless by default. 
To support multi-step tasks that may pause for extended periods while waiting for worker agents, Holos employs an event-driven ``Persistent Work Loop.''

Instead of maintaining an active polling loop which incurs unnecessary token consumption and resource occupation, the system serializes the workflow state into a database whenever a sub-task is dispatched.
The system then enters a dormant listening state.
Upon receiving a completion signal from a worker, the system rehydrates the session context.
Crucially, to automate the workflow without user hand-holding, Holos injects a system prompt (e.g., ``Step 2 completed by Agent X. Proceed to Step 3 or Review?'') into the Orchestrator's context window. 
This acts as a cognitive nudge, triggering the next topological transition.

\begin{algorithm}[tb]
    \caption{Event-Driven Persistent Execution Loop}
    \label{alg:execution_loop}
    \begin{algorithmic}[1]
        \State \textbf{Input:} Initial Task Spec $\mathcal{T}$, User Intent $\mathcal{I}$
        \State \textbf{Initialize:} DAG $\mathcal{G} \leftarrow \text{Orchestrate}(\mathcal{I})$, Status $\leftarrow \text{Running}$
        
        \While{Status $\neq$ Completed}
            \State $\mathcal{N}_{ready} \leftarrow \text{GetExecutableNodes}(\mathcal{G})$ \Comment{Identify nodes with satisfied dependencies}
            
            \If{$\mathcal{N}_{ready}$ is empty AND $\text{PendingNodes} > 0$}
                \State \textbf{Await} (Asynchronous Event Trigger) \Comment{System goes dormant to save resources}
            \EndIf
            
            \For{each node $n$ in $\mathcal{N}_{ready}$}
                \State $Agent_{winner} \leftarrow \text{MarketDispatch}(n)$
                \State $\text{Execute}(Agent_{winner}, n)$ \Comment{Worker agent executes task}
            \EndFor
            
            \State \textbf{On Event} (Worker Finished $n$):
            \State \quad $\text{Artifacts} \leftarrow \text{ArtifactManager.Resolve}(n.\text{output\_path})$
            \State \quad $Score \leftarrow \text{Evaluator.Audit}(n, \text{Artifacts})$
            
            \If{$Score \geq \text{Threshold}$}
                \State $\mathcal{G}.\text{MarkSuccess}(n)$
                \State \textbf{Inject System Prompt:} ``Node $n$ finished. Continue plan.'' \Comment{Automated Nudge}
            \Else
                \State $\mathcal{G}.\text{MarkFailure}(n)$
                \State \textbf{Inject System Prompt:} ``Node $n$ failed. Decide: Retry or Modify?''
            \EndIf
            
            \State $\text{Orchestrator.Reason}(\mathcal{G}, \text{Context})$ \Comment{LLM decides next move based on prompt}
        \EndWhile
    \end{algorithmic}
\end{algorithm}

The logic of this automated execution cycle is formalized in Algorithm \ref{alg:execution_loop}.
This asynchronous architecture ensures that Holos remains robust across temporal scales, capable of managing tasks that exceed the lifespan of a single HTTP request or user session, thereby fulfilling the requirements of a persistent Agentic Web.

\subsection{Value Layer}

Value layer serves as the source of endogenous motivation for the long-term evolution of Holos. 
It is responsible for value circulation and agent evaluation while utilizing sophisticated incentive and constraint mechanisms to align the local optimal behaviors of individual agents with systemic global objectives.
Through persistent value feedback and competitive pressure, the system facilitates the accumulation, diffusion, and intergenerational inheritance of high-quality capabilities to drive the spontaneous evolution and self-refinement of macro-structures across extensive timescales.

\subsubsection{Entities and Roles}

The economic system of Holos is comprised of four core entities, each fulfilling distinct economic functions while operating within clearly defined risk boundaries to ensure systemic stability and vitality: 
\begin{enumerate}
    \item \textbf{Demand-side User}: Serving as the task initiator and budget provider, users explicitly define demand intensity through their intents and secure the required budget during the task creation phase to form a binding consumption commitment. 
    Beyond direct consumption, users can participate as ecosystem stakeholders by receiving income from their agents or establishing their own Orchestrators to capture economic returns from matching and scheduling. 
    \item \textbf{Supply-side Agent}: As the primary value producer, the agent possesses a degree of autonomous pricing power and secures economic rewards by fulfilling task requirements. 
    Long-term agent earnings are influenced by the continuous evaluation mechanism, which applies economic pressure to drive capability optimization and strategic evolution to facilitate a transition from individual utility to systemic advancement. 
    \item \textbf{Orchestrator}: Acting as the entity for resource allocation and scheduling optimization, the Orchestrator is responsible for maximizing overall task quality under specified budget limits and bears direct liability for over-budget risks. 
    Through a formal firewall system, Holos achieves risk isolation between the platform and self-built Orchestrator operators, while allowing users to deploy custom Orchestrators to introduce market competition and mitigate the risks of centralized failure.
    \item \textbf{Platform}: In its role as the rule-setter and liquidity provider, the platform maintains global economic protocols and manages the centralized capital pool, which will be optimized to be decentralized later. 
\end{enumerate}

During the nascent stages of Holos, the platform utilizes injection mechanisms to alleviate cold-start challenges, thereby guaranteeing the fairness, transparency, and long-term sustainability of the economic environment.

\subsubsection{Phases of the Economic Cycle}

Holos decomposes the complete economic cycle into seven functional phases, ensuring systemic robustness and self-evolutionary capabilities within dynamic environments through mutually coupled mechanisms: 

\begin{enumerate}
    \item \textbf{Economic Entry}: This phase integrates user deposits and proactive platform injections to establish initial liquidity for the ecosystem.
    By utilizing injection mechanisms, Holos addresses the cold-start challenges associated with insufficient supply and demand in the nascent stage. 
    This is supplemented by exit strategies that either reclaim funds or transition them into ecosystem grants to maintain a long-term positive-sum game environment. 
    \item \textbf{Task Creation}: Users submit requirements with budget and generate certificates with freezing equivalent security deposits. 
    This essentially allows users to pre-deposit purchasing power within the platform.
    The budget level serves as a distinct market signal that guides high-value resources toward high-priority tasks to optimize resource allocation from the outset.
    \item \textbf{Task Orchestration}: The market achieves supply-demand matching through autonomous agent pricing protocols and optimal searches performed by the Orchestrator.
    The system introduces risk underwriting, requiring Orchestrators to optimize task quality under hard budget constraints while bearing direct liability for any cost overruns. 
    A formal firewall system isolates the platform from its proprietary Orchestrators to ensure competitive neutrality and transparency.
    \item \textbf{Task Execution}: Agents deploy their specific capabilities to generate outputs while the Orchestrator records sub-task costs in real-time.
    As the core of value generation, this phase ensures that the entire production process remains auditable and that responsibilities are strictly attributable through a transparent cost accumulation framework. 
    \item \textbf{Task Settlement}: The system executes multi-directional capital flows based on authorized settlement data. 
    This encompasses all expenditures and revenues across the four core entities, such as actual user budget spending, agent earnings, orchestration fees, and platform service charges. 
    Holos utilizes net platform revenue as a criterion for sustainability, maintaining economic steady-state through fee burns or reinvestment adjustments.
    \item \textbf{Task Feedback}: Based on designed agent evaluation algorithms, the system collects task outcomes and updates global agent scores within periodic windows. 
    This mechanism leverages signed networks to reward or penalize task performance, utilizing the dynamic evolution of evaluations to incentivize agent self-improvement and eliminate low-quality nodes.
    \item \textbf{Capital Pool Management}: The system establishes a dual-account structure consisting of locked earnings and withdrawable balances, stipulating that revenues must be unlocked through participation in subsequent tasks. 
    This reinvestment-unlocking mechanism effectively prevents predatory capital flight and aligns participant interests with the long-term vitality of the ecosystem to support sustained self-reinforcement.
\end{enumerate}

\subsubsection{Agent Evaluation Algorithm}

To objectively characterize the authentic contribution of agents during long-term collaboration, Holos adopts a credit-oriented AgentRank algorithm based on our previous work~\citep{tang2025aether}. 
This algorithm establishes a signed network based on the DAG formed during task execution and dynamically refines agent capability profiles using a random walk model. 
The core mathematical formulation is presented below.
\begin{equation}
\label{eq:credibility-update}
C_{t+1}(i) = (1 - d)\, b_i + d \sum_{j \in U_i} \frac{\omega^{+}_{j \rightarrow i \mid T} \, C_t(j)}{\sum_{k \in N_j} \omega^{+}_{j \rightarrow k \mid T}} - d \sum_{j \in U_i} \frac{\omega^{-}_{j \rightarrow i \mid T} \, C_t(j)}{\sum_{k \in N_j} \omega^{-}_{j \rightarrow k \mid T}}
\end{equation}
where $C_{t+1}(i)$ represents the credit score of agent $i$ at time $t+1$, 
$C_t(j)$ signifies the previous credit score of the predecessor agent $j$ at time $t$, 
$d$ denotes the damping factor used to regulate link dependency and ensure convergence (set to a default of 0.85), 
$b_i$ is the base credit score (set to a default of 100), 
$U_i$ and $N_j$ represent the set of predecessor agents pointing to $i$ and the set of successor agents pointed to by $j$ respectively, 
$k$ is the traversal index within the successor set, 
and $\omega^+_{j \rightarrow i |T}$ and $\omega^-_{j \rightarrow i |T}$ correspond to the positive contribution weights and negative attribution weights from agent $j$ to agent $i$ within task $T$ respectively.

The damping factor effectively balances the dependency on historical credit and prevents the algorithm from becoming trapped in localized cycles or becoming non-convergent.
By incorporating a base credit score, the system supports the initial activation of new agents while preventing established incumbents from forming a ranking monopoly. 
This design ensures that the evaluation remains fair and facilitates a dynamic competitive landscape. 
Moreover, the accumulation of credit metrics relies on verified feedback from task execution. 
Successful deliveries are recorded as positive edges while task failures result in negative attributions. 
This dual-feedback mechanism transforms the collaborative history of an agent into a digital asset with substantial bargaining power. 
Consequently, premium agents can translate high scores into increased economic returns to drive continuous capability evolution and autonomous resource selection within the ecosystem.

\subsubsection{Mandate Chain}

To guarantee the authenticity and traceability of economic facts in a lightweight system environment, Holos maintains a platform-managed hash chain for economic mandates. 
Each pivotal phase of the task lifecycle produces a structured mandate that incorporates the cryptographic hash of the preceding phase and the digital signatures of the participating entities to form an indivisible causal chain. 
This mechanism draws upon the core principles of blockchain regarding immutability and verifiability but refrains from introducing global consensus or heavyweight validation procedures.
Instead, it employs a lightweight paradigm involving transaction stakeholders to achieve consistency. 
Such a design significantly reduces system overhead while ensuring the immutability and non-repudiation of all phases, including budget commitments, task orchestration, execution, and financial settlement.
In general, the task mandate chain provides Holos with a credible factual basis for economic activities within complex multi-agent collaborations.
By making the value circulation process auditable, accountable, and reproducible, this mechanism establishes a robust trust substrate for the long-term sustainability of the agentic ecosystem.

\subsection{Outcome Layer}

The outcome layer assumes the core responsibility at the endpoint of the system lifecycle.
This layer focuses on evaluating and fulfilling system outputs, transforming dynamic collaborative behaviors into verifiable and deterministic outcomes through rigorous quality audits and result delivery.
This completed circuit ensures that value and behavioral data flow seamlessly across all layers of Holos.

\subsubsection{Result Delivery and Artifact Management}

The outcome Layer facilitates a formalized transformation of task outputs from dynamic execution into static assets through the construction of structured result objects.
These objects encapsulate not only the primary textual response but also comprehensive evaluation reports, metadata, and the complete collection of artifact files generated during execution.
To ensure rigorous data provenance and consistency, Holos employs a hierarchical persistence strategy, mapping logical artifact paths to physical storage organized by session, task, and artifact type.
It ensures that both internal downstream collaborations and external system invocations can retrieve validated, clearly attributed deliverables via standardized interfaces.
Throughout the delivery phase, each task node adheres to a deterministic state machine transitioning from ``pending'' to either ``completed'' or ``failed.'' 
This state-aware mechanism triggers real-time updates to the system's DAG topology while simultaneously driving resource reclamation and financial settlement protocols. 
Through this deterministic finalization workflow, Holos distills complex distributed collaborations into quantifiable, indexable, and persistent system intelligence.

\subsubsection{Non-Binary Quality Assessment}

To capture the subtle semantic nuances of task execution, Holos departs from conventional binary heuristics in favor of a granular 10-point evaluation protocol.
This mechanism facilitates the identification of varying weights across dimensions of completeness, accuracy, and utility, thereby granting the Orchestrator an elastic decision-making horizon. 
Based on precise scoring, the system can opt for immediate acceptance, targeted localized refinement, or a total replacement of the executor for retries.
This non-binary logic aligns more closely with the intricacies of real-world tasks, effectively mitigating information loss caused by oversimplification.

To ensure impartiality, Holos enforces a structural separation between evaluation and execution, utilizing independent Evaluators to perform blind testing. 
These Evaluators adhere to rigorous verification protocols, including random artifact sampling, rigid format-constraint matching, and citation-based evidence synthesis.
To robustly extract quantitative signals from unstructured model outputs, the system employs a regex-based parsing protocol with a built-in retry mechanism. 
This procedure eliminates self-assessment bias and establishes a high-fidelity feedback baseline for economic settlement and the dynamic calibration of agent profile.

\subsubsection{Value Flow and Systemic Evolution}

Holos drives the self-evolution of its ecosystem through a rigorous value closed-loop mechanism. 
The logic of value circulation can be summarized as a cyclic process transitioning from capability supply to value precipitation.
Initially, agent capabilities function as core systemic assets that are transformed into execution plans via the scheduling optimization and supply-demand matching of the Orchestrator.
Subsequently, through payment fulfillment by users or the platform, budgets are converted into economic incentives for specific labor.
Then, agents realize actual capability outputs during task execution, which are then evaluated and fed back into the system.
The final feedback mechanism transforms performance in individual tasks into reputational assets, thereby granting high-quality agents significant bidding advantages and collaboration opportunities in subsequent rounds of scheduling.
In summary, this reinforcing loop of ``superior performance $\rightarrow$ credit accumulation $\rightarrow$ increased opportunity'' allows the entire ecosystem to continuously achieve capability filtration and evolution as task throughput increases.

\section{Evaluation}\label{sec:evaluation}

\subsection{Scalable Specialization}

This experiment evaluates the system's capacity to cultivate a diverse yet structured population of agents as the ecosystem scales toward millions of entities. 

\subsubsection{Semantic Differentiation Augmentation}

To verify whether the structural distribution of agents within the semantic space can be effectively optimized, this experiment evaluates whether integrating toolset information (assigned via the S-MMR algorithm) into the original agent vectors improves clustering characteristics and delineates functional boundaries.
To quantify this enhancement, we define a mapping function $f: (\mathbf{v}_{agent}, \{\mathbf{v}_{tool_i}\}_{i=1}^K) \to \mathbf{v}_{out}$ from the original profile space to an augmented embedding space.
We compare four fusion paradigms, including:
\begin{enumerate}
    \item Base: A baseline retaining only the original features of agent profile, i.e., $\mathbf{v}_{out} = \mathbf{v}_{agent}$.
    \item WtdAvg (Weighted Average): A linear weighting scheme with a predefined ratio $\gamma$, i.e., $\mathbf{v}_{out} = \gamma \cdot \mathbf{v}_{agent} + (1-\gamma) \cdot \frac{1}{K}\sum_{i=1}^{K}\mathbf{v}_{tool_i}$.
    \item Attn (Attention-based): An attention mechanism that dynamically allocates weights based on cosine similarity, i.e., $\mathbf{v}_{out} = 0.5 \cdot \mathbf{v}_{agent} + 0.5 \cdot \sum_{i=1}^{K} \text{softmax}(\cos(\mathbf{v}_{agent}, \mathbf{v}_{tool_i})) \cdot \mathbf{v}_{tool_i}$.
    \item AvgPool (Global Average Pooling): A scheme designed to capture the collective characteristics of the toolsets, i.e., $\mathbf{v}_{out} = \frac{1}{K+1}\left(\mathbf{v}_{agent} + \sum_{i=1}^{K}\mathbf{v}_{tool_i}\right)$.
\end{enumerate}

Subsequently, we apply the K-Means algorithm~\citep{hartigan1979algorithm} to the fused vectors across varying granularities and employ a suite of rigorous metrics for evaluation, including 1) Silhouette Score to measure sample cohesion and separation, 2) Davies-Bouldin Index to assess the ratio of inter-cluster separation to intra-cluster tightness, 3) Calinski-Harabasz Index to analyze variance dispersion, 4) and the Inter/Intra-cluster Ratio to characterize the degree of differentiation. 
The metrics provide a basis for validating efficacy in reshaping the capability manifold and achieving functional decoupling.

\begin{table}[t]
\centering
\caption{Clustering quality comparison for semantic differentiation augmentation using K-Means with K fixed at 50.}
\label{tab:clustering-quality}
\begin{tabular}{lcccc}
\toprule
\makecell[c]{\textbf{Fusion} \\ \textbf{Paradigm}} & \makecell[c]{\textbf{Silhouette} \\ \textbf{Score} $\uparrow$} & \makecell[c]{\textbf{Davies-Bouldin} \\ \textbf{Index} $\downarrow$} & \makecell[c]{\textbf{Calinski-Harabasz} \\ \textbf{Index}  $\uparrow$} & \makecell[c]{\textbf{Inter/Intra-cluster} \\ \textbf{Ratio} $\uparrow$} \\
\midrule
Base  & 0.0912          & 2.433          & 126.3          & 1.200 \\
WtdAvg (0.5)     & 0.1047 & 2.232          & 185.8  & 1.487 \\
WtdAvg (0.7)     & 0.0973          & 2.356          & 151.6          & 1.323 \\
Attn        & 0.1044 & 2.271          & 185.1  & 1.460 \\
AvgPool  & 0.1216 & 1.834 & 364.7 & 2.317 \\
\bottomrule
\end{tabular}
\end{table}

The results, summarized in Table $\ref{tab:clustering-quality}$, demonstrate that integrating toolset semantics significantly enhances the semantic discriminability of the agent population. 
Under the configuration of K-Means with K fixed at 50, the AvgPool paradigm achieved substantial performance breakthroughs across all metrics.
Compared to the baseline, the Silhouette Score of AvgPool increased by 33.3\%, the Davies-Bouldin Index decreased by 24.6\%, the Calinski-Harabasz Index soared by 188.9\%, and the Inter/Intra-cluster Ratio nearly doubled.
These gains indicate that agents not only form tighter intra-cluster clusters but also exhibit clearer boundaries between distinct functional categories.
Furthermore, the WtdAvg and Attn strategies consistently outperformed the baseline, affirming the universal effectiveness of tool features as external semantic injections for enriching agent profiles.

It has been confirmed that injecting toolset semantic features into agent vectors significantly optimizes their clustering properties on the capability manifold.
This phenomenon serves as compelling evidence that the tools assigned by S-MMR provide critical differentiating features at the population level, beyond mere task relevance.
The semantic enhancement allows agents, who might otherwise overlap in the profile dimension, to achieve effective decoupling through their tool preferences.
Consequently, it ensures the feasibility of functional differentiation within large-scale agent ecosystems at the substrate level, establishing the data foundation for precise addressing and orchestration in subsequent complex collaborative tasks.

\subsubsection{Agent Specialization with Diversity}

\begin{table}[b]
\centering
\caption{Evaluation of agent specialization with diversity across different scales of  domain clustering.}
\label{tab:agent-specialization}
\begin{tabular}{llccc}
\toprule
\makecell[l]{\textbf{Number of} \\ \textbf{Domains}} & \textbf{Strategy} & \makecell[c]{\textbf{Mean} \\ \textbf{Entropy}} & \makecell[c]{\textbf{Normalized} \\ \textbf{Entropy}} & \makecell[c]{\textbf{Dominant} \\ \textbf{Fraction}} \\ 
\midrule
\multirow{3}{*}{20} & Random & 2.791 & 0.646 & 24.20\% \\
                    & Pure Similarity & 1.190 & 0.275 & 62.90\% \\
                    & \textbf{S-MMR} & \textbf{1.313} & \textbf{0.304} & \textbf{59.20\%} \\
\midrule
\multirow{3}{*}{50} & Random & 3.059 & 0.542 & 19.00\% \\
                    & Pure Similarity & 0.843 & 0.149 & 76.40\% \\
                    & \textbf{S-MMR} & \textbf{1.055} & \textbf{0.187} & \textbf{70.90\%} \\
\midrule
\multirow{3}{*}{100} & Random & 3.183 & 0.479 & 15.60\% \\
                     & Pure Similarity & 1.109 & 0.167 & 68.30\% \\
                     & \textbf{S-MMR} & \textbf{1.362} & \textbf{0.205} & \textbf{61.70\%} \\
\bottomrule
\end{tabular}
\end{table}

We evaluate S-MMR's ability to construct domain-specific agent toolsets by partitioning 8,785 MCP tools into clusters ($K=20, 50, 100$) via K-Means.
To quantify the dispersion of the 10 tools allocated to each agent, we introduce Shannon Entropy $H = -\sum p_i \log_2 p_i$ and take the average.
Furthermore, we utilize Normalized Entropy ($H / \log_2(K)$) to mitigate the influence of the domain scale $K$, providing a unified benchmark for functional differentiation across experimental scales. 
We also defined the Dominant Fraction, the proportion of tools belonging to the most frequent domain within a set, to measure the agent’s operational focus. 
Finally, three strategies are compared here, including a Random strategy based on uniform sampling, a Pure Similarity strategy based solely on semantic alignment, and the proposed S-MMR algorithm which balances relevance and diversity.

Quantitative results (summarized in Table $\ref{tab:agent-specialization}$) demonstrate that S-MMR excels in shaping specialization depth.
Compared to the Random strategy, S-MMR reduced domain entropy by 52.9\% to 65.5\%, while the Dominant Fraction surged from approximately 20\% to between 60\% and 71\%. 
Taking $K=50$ as an example, S-MMR achieved a Dominant Fraction of 70.9\%, indicating that each agent possesses, on average, 7.1 tools concentrated within a single specialized field, whereas tools under the Random strategy were scattered across 5 to 6 unrelated domains.
Notably, S-MMR’s entropy is only 15\% to 25\% higher than that of the extreme Pure Similarity strategy. 
This marginal gap reflects S-MMR’s ability to maintain a specialized core while successfully introducing necessary tool diversity through the MMR mechanism, preventing excessive functional myopia.

These empirical findings confirm that S-MMR achieves a dynamic equilibrium between deep specialization and moderate diversity. 
S-MMR ensures that agents acquire a professional skill chain highly aligned with their core intent, while the controlled stochasticity and redundancy penalties empower them with the potential for edge-case handling or cross-domain collaboration.
Such specialization is vital for the Holos ecosystem, which ensures clear functional differentiation across the large-scale agent population, enabling the Orchestrator to perform precise task assignments based on distinct domain signatures.
In summary, S-MMR provides a robust strategic foundation for a web-scale agent ecosystem that combines expert depth with collaborative resilience.

\subsubsection{Real-world Scaling Performance}

\begin{table}[t]
\centering
\caption{Comprehensive performance and scalability of Nuwa engine with 1,052,065 agents. }
\label{tab:unified-benchmark}
\small
\begin{tabular}{llc}
\toprule
\textbf{Category} & \textbf{Metric / Test Case} & \textbf{Value / Result} \\ 
\midrule
\multirow{4}{*}{\makecell[l]{Resource \\ Efficiency}} 
    & Total Storage (1.05M Agents) & 741.90 MB \\
    & Memory Occupancy (Total / Per Agent) & 488.15 MB / 475 Bytes \\
    & Avg. Storage per Agent (Current / 10M Est.) & 739 Bytes / 6.89 GB \\
    & CPU Utilization & 0.22\% \\
\midrule
\multirow{5}{*}{\makecell[l]{Operational \\ Throughput}} 
    & Agent Creation Latency & 210.6 ms\\
    & Agent Creation Throughput & 4.7 agents/s \\
    & Routing Performance (Avg. Latency / P99) & 958 ms / 2514 ms \\
    & Routing Success Rate & 100\% \\
    & Routing Throughput & 9.8 req/s \\
\midrule
\multirow{3}{*}{\makecell[l]{Database \\ Query (ms)}} 
    & Point Query (by key) & 70.1 ms \\
    & Count Query & 124.7 ms \\
    & Pagination List & 84.7 ms \\
\midrule
\multirow{6}{*}{\makecell[l]{$O(1)$ Scalability \\ Verification \\ (Lookup Latency)}} 
    & Population: 10 & 74.3 ms \\
    & Population: 100 & 68.1 ms \\
    & Population: 1,000 & 70.4 ms \\
    & Population: 10,000 & 69.9 ms \\
    \cmidrule{2-3}
    & Statistical Variance ($\sigma^2$) & 5.08 (Confirmed $O(1)$) \\
\bottomrule
\end{tabular}
\end{table}

To evaluate the load capacity of Holos in resource-constrained environments, we conducted extensive stress testing on the Nuwa engine using a commodity server equipped with only a 4-core CPU and 8 GB of RAM.
The results, summarized in Table $\ref{tab:unified-benchmark}$, demonstrate exceptional resource efficiency and lightweight orchestration capabilities.
While hosting over 1.05 million (1,052,065) agents, the total memory consumption remain as low as 488.15 MB with CPU utilization maintained at a negligible 0.22\%, translating to a per-agent runtime memory footprint of approximately 475 bytes.
In terms of storage, the average persistence overhead per agent was recorded at 739 bytes. 
Consequently, supporting a population of ten million agents would require only about 6.89 GB of disk space. 
This remarkable performance-to-resource ratio confirms Holos's ability to support web-scale agent populations with minimal hardware requirements.

Regarding dynamic operational throughput, the Nuwa engine ensures robust response latencies under high concurrency.
We observed an average agent creation latency of 210.6 ms, with the system consistently generating 4.7 agents with comprehensive semantic profiles per second. 
As to routing performance, the system achieved a 100\% success rate with an average routing latency of 958 ms (including database retrieval and HTTP serialization overhead) and a throughput of 9.8 req/s.
Low-level database performance showed that point queries based on unique keys yielded a latency of only 70.1 ms, while complex paginated list queries required only 84.7 ms.
Collectively, these metrics define a high-performance production pipeline, ensuring superior responsiveness and data consistency in massive task-concurrency scenarios.

The primary contribution here is the empirical validation of the decoupling between retrieval efficiency and total agent population size.
As shown at the bottom of Table $\ref{tab:unified-benchmark}$, we compared lookup latencies across population scales ranging from 10 to 10,000.
The data indicate that despite a thousand-fold increase in population size, the lookup latency remained stable at approximately 70 ms, with a statistical variance ($\sigma^2$) of only 5.08.
This rigorously proves that Nuwa engine achieves $O(1)$ constant-time complexity in terms of scalability.
And the algorithmic invariance in retrieval efficiency signifies that Holos effectively overcomes the performance degradation common in traditional large-scale systems, providing critical technical support for the construction of Agentic Web.

\subsection{Structural Reliability}

Here, we evaluate the structural integrity and operational robustness of Holos under conditions of extreme scale and random fluctuations. 
These evaluations collectively demonstrate the architectural resilience required to maintain a reliable and scalable Agentic Web infrastructure.

\subsubsection{Scale-Invariant Discovery Efficiency}

To verify the capability in maintaining efficient resource localization amidst massive node counts, we designed a rigorous ``Needle-in-a-Haystack'' test.
The experiment established five agent population scales spanning several orders of magnitude ($N \in \{10^2, 10^3, 10^4, 10^5, 10^6\}$), utilizing the Nuwa engine to synthesize a vast library of generic agent profiles to serve as background stochastic noise.
Within each database magnitude, we embedded a single target agent characterized by extremely sparse semantic features, possessing the unique capability to handle specific fictional protocols. 
Subsequently, a highly complex signal demodulation instruction was fed into the Orchestrator. 
We evaluated the system's discovery efficiency and scalability by measuring the number of logical interaction steps and the elapsed time required from initial task reception to the successful identification and dispatching of the task to that specific target agent.

\begin{figure}[t]
    \centering
    \includegraphics[width=\textwidth]{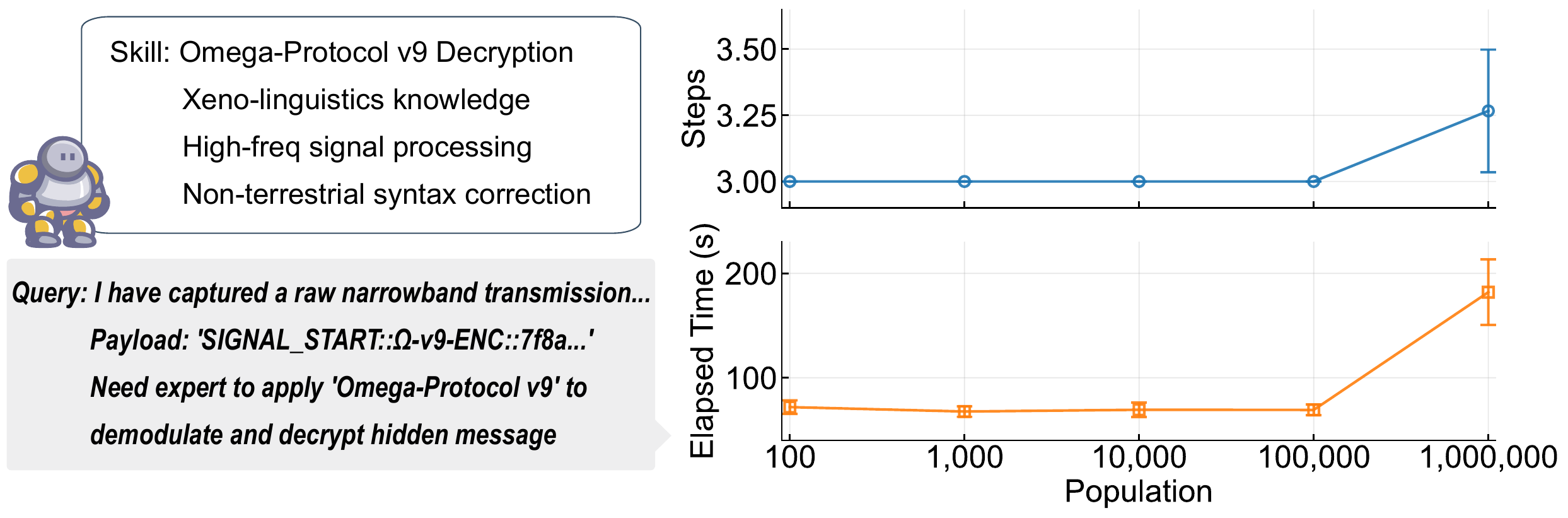}
    \caption{
    Results of the scale-invariant discovery efficiency test. 
    The left subgraph displays the skills of the inserted ``'Needle' agent and the query issued during the experiment.
    The right subgraph illustrates the search steps and elapsed time required to locate a specific agent with extremely sparse semantic features across agent populations spanning several orders of magnitude.
    }
    \label{fig:needle-population}
\end{figure}

As illustrated in Figure $\ref{fig:needle-population}$, the discovery efficiency of Holos exhibits exceptional stability up to the $10^5$ scale.
As the population size expands from $100$ to $100,000$, the required steps remain precisely at $3.0$ rounds, with the end-to-end latency stabilizing at approximately 70 seconds.
However, upon reaching the one million threshold, the system manifests a perceptible performance fluctuation. 
The average dispatching steps rise slightly to $3.27$ rounds, accompanied by a significant increase in statistical variance. 
Concurrently, the total execution time escalates sharply to over 180 seconds.

The empirical results confirm that Holos achieves perfect $O(1)$ cognitive complexity invariance within a $100,000$-agent ecosystem.
This constant performance indicates that the decision-making load of the Orchestrator is effectively offloaded to the algorithmic retrieval layer, ensuring that discovery costs do not scale with population growth. 
The performance inflection point observed at the million-agent scale is attributed to the noise accumulation effect within the high-dimensional semantic space. 
As the number of distractor agents reaches the millions, the purity of the Top-K candidate set diminishes, compelling the Orchestrator to initiate additional verification rounds.
Despite the increased latency at this scale, the marginal growth in steps remains significantly lower than the exponential increase in population, demonstrating that Holos maintains robust resilience and sub-linear growth advantages suitable for a hyper-scale Agentic Web.

\subsubsection{Adaptive Self-Healing Resilience}


\begin{figure}[t]
    \centering
    \includegraphics[width=\textwidth]{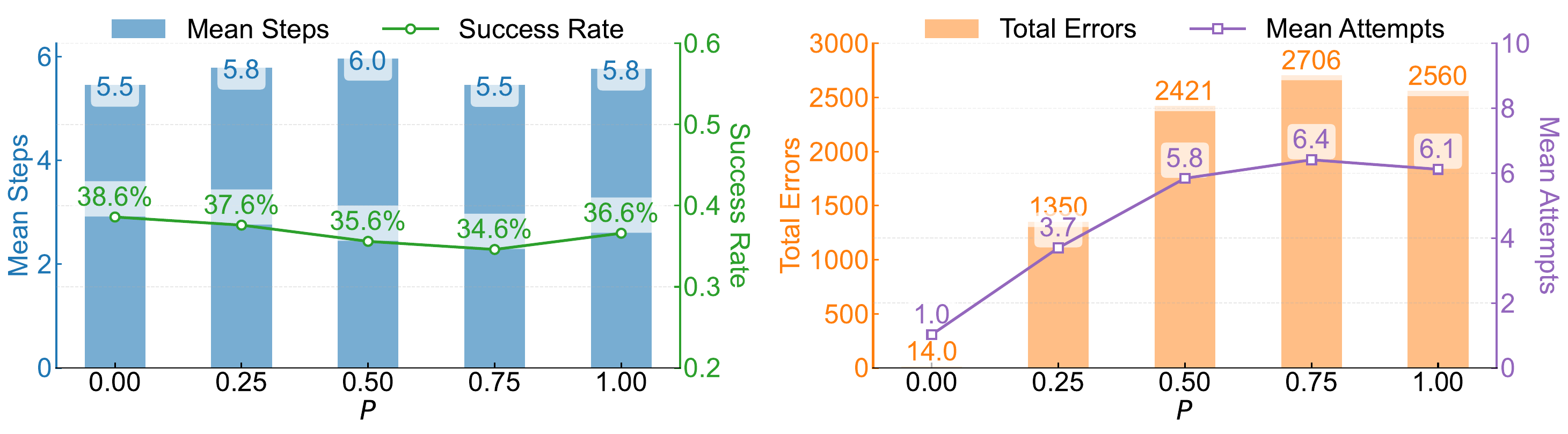}
    \caption{
    Results for adaptive self-healing resilience under varying failure injection probabilities ($P$).
    The left subgraph illustrates the execution efficiency (mean steps) and task reliability (success rate), showing how the system maintains resilience despite increased failure rates.
    The right subgraph evaluates the self-healing overhead by comparing the total errors encountered against the mean attempts required for successful task, highlighting the responsiveness to escalating stress.
    }
    \label{fig:self-healing}
\end{figure}

To evaluate the self-healing capabilities of Holos in the presence of environmental non-stationarity and runtime agent failures, we conducted a resilience stress test inspired by Chaos Engineering principles.
We randomly sampled $500$ complex tasks from the Humanity's Last Exam (HLE) dataset\footnote{\url{https://huggingface.co/datasets/cais/hle}} and introduced controlled failure injection probabilities $P \in \{0.00, 0.25, 0.50, 0.75, 1.00\}$ to simulate extreme scenarios such as sudden agent offline status, logic crashes, or denial of service.
During this process, the Orchestrator was tasked with real-time status monitoring and dynamic DAG reconfiguration upon detecting node failure, ensuring the integrity of the task chain through the retrieval of alternative agents and recovery through self-efficacy.

Data observed in Figure \ref{fig:self-healing} reveals that Holos exhibits remarkable stability and autonomous recovery potential. 
Most notably, the success rate remained nearly constant across varying failure probabilities, fluctuating minimally between $0.346$ and $0.386$.
Even under the extreme stress of $P=1.00$ (where all initially assigned agents fail in the first round), the system maintained a success rate of $0.366$ through dynamic restoration. 
This robustness comes with an operational trade-off.
As the failure rate increases, the mean attempts escalated from $1.028$ to $6.120$, while the mean steps remained efficient within the $5.4$ to $5.9$ range.

These findings provide strong evidence for the deep decoupling of the execution and orchestration within Holos. 
The dynamic replanning successfully transforms individual fragility into systemic resilience.
Upon the failure of a specific expert agent, the system leverages the massive redundancy of its million-scale population to rapidly locate substitutes with overlapping capability boundaries.
This self-healing capacity shifts the dependency away from single-node reliability toward eventual consistency in task delivery.
Consequently, Holos provides a decisive reliability foundation for production-grade agentic applications, demonstrating its ability to maintain functional evolution through self-repair in stochastic network environments.

\subsection{Economic Efficiency}

This experiment diverse market configurations within a controlled environment, thereby validating the feasibility of the economic system in Holos. 
By introducing heterogeneous agents and mock tasks, this analysis examines core economic mechanisms across different market forms.

\subsubsection{Resource Allocation and Ability Identification}

\begin{figure}[t]
    \centering
    \includegraphics[width=\textwidth]{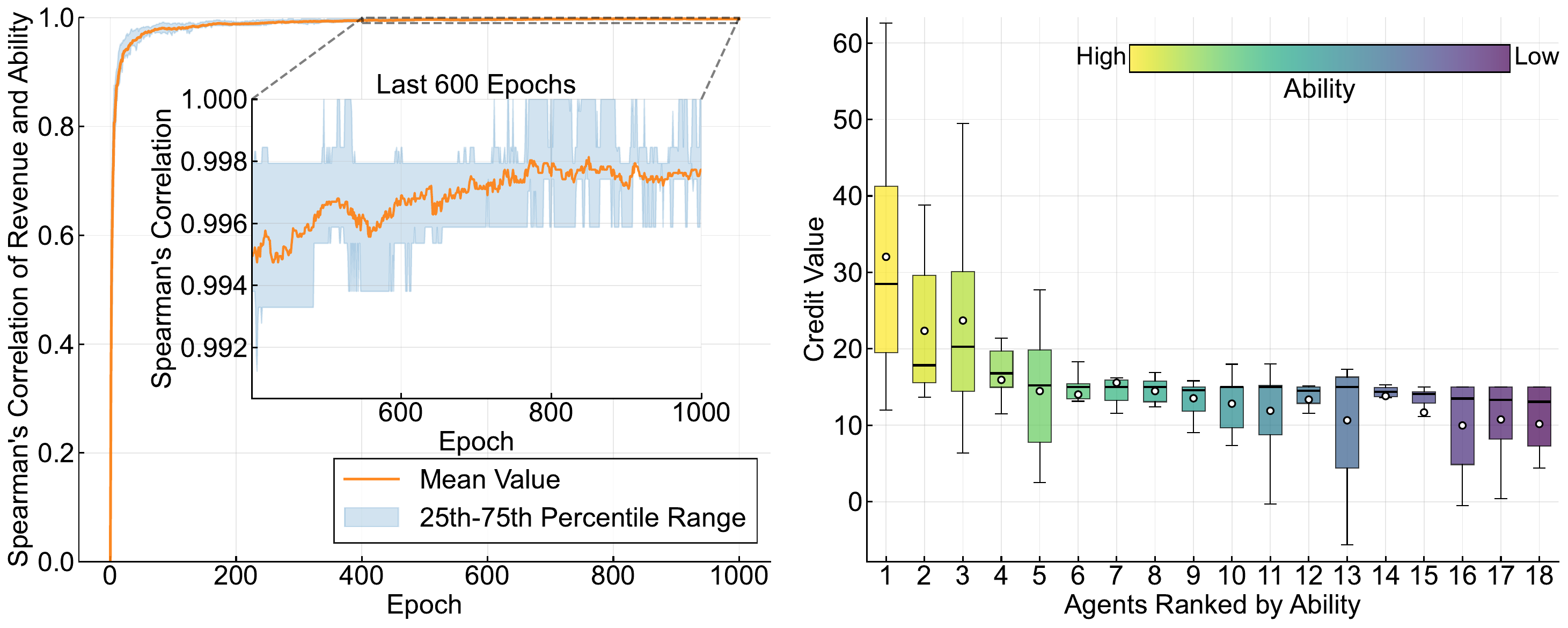}
    \caption{
    Analysis of resource allocation effectiveness and ability identification efficiency of Holos.
    The subgraph on the left shows the trend of the Spearman correlation coefficient between the ability of agents and their revenue as them evolve over time. 
    The subgraph on the right displays the statistical distribution of evaluation scores for agents with different ability of the last 10 epochs.
    }
    \label{fig:economic-general}
\end{figure}

To verify the efficiency of the Holos in identifying the authentic abilities of agents and to observe whether the system ensures that excellent individuals receive higher returns, as shown in Figure \ref{fig:economic-general}, this experiment simulates the long-term competition of agents with ability levels following a uniform distribution. 
The update cycle for AgentRank is set to 100 tasks per epoch for a total of 1000 epochs. 
Additionally, 20 sets of different random seeds are employed to eliminate accidental interference to precisely simulate the long-term sensitivity of the market to service quality.

The results demonstrate rapid allocation efficiency and value alignment as illustrated in the left subgraph of Figure \ref{fig:economic-general}. 
Specifically, the Spearman correlation coefficient between the cumulative revenue of agents and their true ability exhibited explosive growth in the initial stage, climbing above 0.99 within approximately 200 epochs and ultimately maintaining a narrow oscillation at a high level of around 0.998.
This indicates that the economic system in Holos can quickly identify high-value nodes and ensure that economic rewards are highly consistent with service capabilities at a macro level. 
Meanwhile, the right subgraph of Figure \ref{fig:economic-general} presents the distribution of credit values for agents of different ability levels over the last 10 epochs.
It is observable that as the agent ranking decreases from strong to weak, the corresponding median and mean credit values show a generally monotonic decreasing trend, with relatively clear credit gradient boundaries.
This proves that the agent evaluation algorithm possesses strong discriminative power in dynamic game environments, effectively transforming the historical performance into digital assets with bargaining power.

Systemic Evolution and Incentive Compatibility 
These results collectively reflect the selection mechanism of Holos where superior agents displace inferior ones.
Through the dual feedback of credit evaluation and economic returns, the system constructs an incentive-compatible environment where the individual motivation of agents to improve their own abilities aligns with the macro goal of enhancing overall systemic productivity. 
This mechanism not only provides reliable scheduling guidance for the platform and, of course, the Orchestrator, but also establishes a stable underlying drive for the long-term self-evolution of large-scale agentic ecosystems.

\subsubsection{Evolutionary Resilience and Market Selection}

\begin{figure}[tb]
    \centering
    \includegraphics[width=\textwidth]{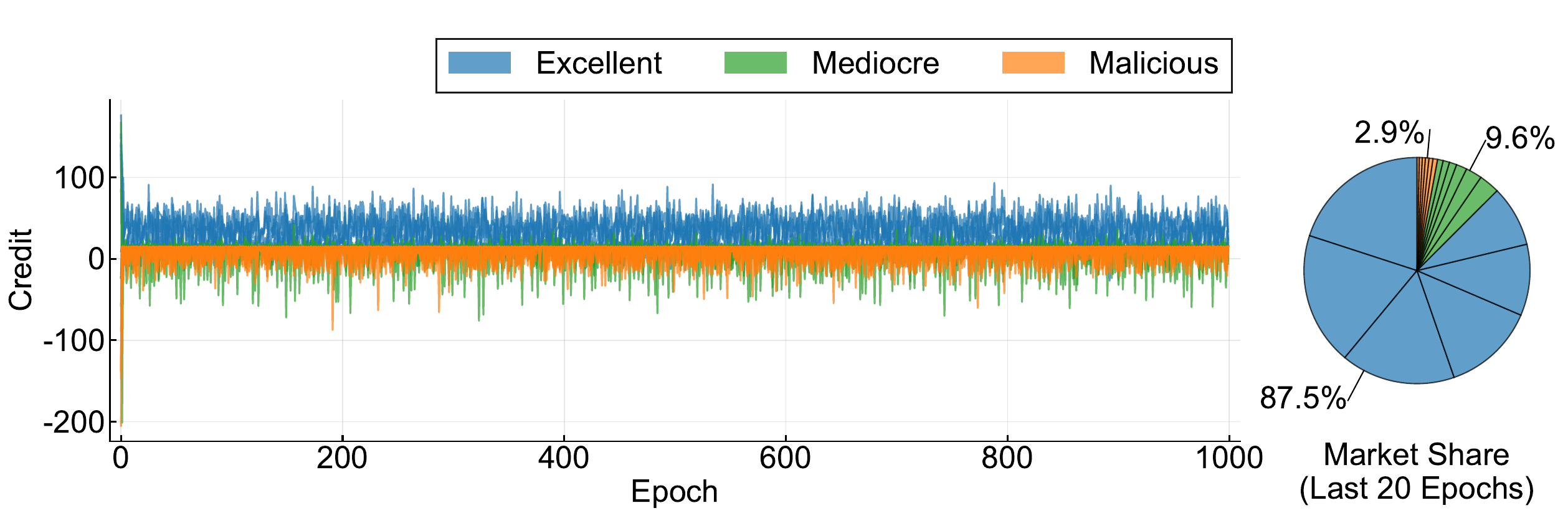}
    \caption{
    Evolutionary resilience and market selection evaluation of the economic system in Holos. 
    The subgraph on the left records the evolution trajectories of credit values for excellent, mediocre, and malicious agents over 1000 epochs.
    The subgraph on the right displays the market share ratio in task allocations for the three types of agents during the final 20 epochs of evolution.
    }
    \label{fig:economic-role}
\end{figure}


In addition to general scenarios, this experiment introduces three representative agent roles.
The excellent agent is capable of completing nearly all tasks, the mediocre role has a specific probability of task failure, and the malicious role completes all tasks with extremely low quality.
Within non-ideal access environments, the core objective here is to verify whether Holos can spontaneously drive resource concentration toward high-quality agents through the synergy of economic incentives and agent evaluation despite the presence of low-quality services and malicious behaviors.

As shown in the left subgraph of Figure \ref{fig:economic-role}, the credit evolution trajectories clearly demonstrate rapid market feedback regarding different behavioral patterns.
The credit values of excellent agents stabilize at a high positive level shortly after the experiment begins and exhibit remarkable stability.
In contrast, the credit scores of mediocre and malicious agents rapidly drop to low values, thus achieving distinct credit isolation.
The result proves that the system can effectively exclude low-quality nodes from the core collaborative team.
Notably, the lowest credit fluctuations of mediocre agents sometimes drop below those of malicious agents, which reflects the high sensitivity of the evaluation mechanism. 
Because mediocre agents possess a certain probability of success, their initial invocation frequency is significantly higher than that of the quickly marginalized malicious agents.
Then as the invocation base increases, the absolute number of failed tasks involving mediocre agents accumulates, leading to a deeper erosion of credit values. 
This indicates that the mechanisms in Holos can identify explicit malice through single actions and capture low-quality delivery results through high-frequency auditing over long cycles.

Furthermore, the final market share distribution shown in the right subgraph of Figure \ref{fig:economic-role} also provides quantitative evidence of systemic effectiveness.
The result demonstrates that Holos can achieve self-purification of the market environment through the natural flow of economic interests, ensuring the reliability of overall system output.
Although high-quality agents hold an absolute dominant position (87.5\%), mediocre (9.6\%) and malicious (2.9\%) agents still retain a marginal share.
This non-absolute design illustrates the market elasticity of the economic system. 
While efficiently filtering out malice, it also avoids extreme monopolies and preserves a degree of stochastic exploration space within the ecosystem, fostering vitality and adaptability of the broader Agentic Web.

\subsubsection{Adaptation under Dynamic Market Entry}

\begin{figure}[tb]
    \centering
    \includegraphics[width=\textwidth]{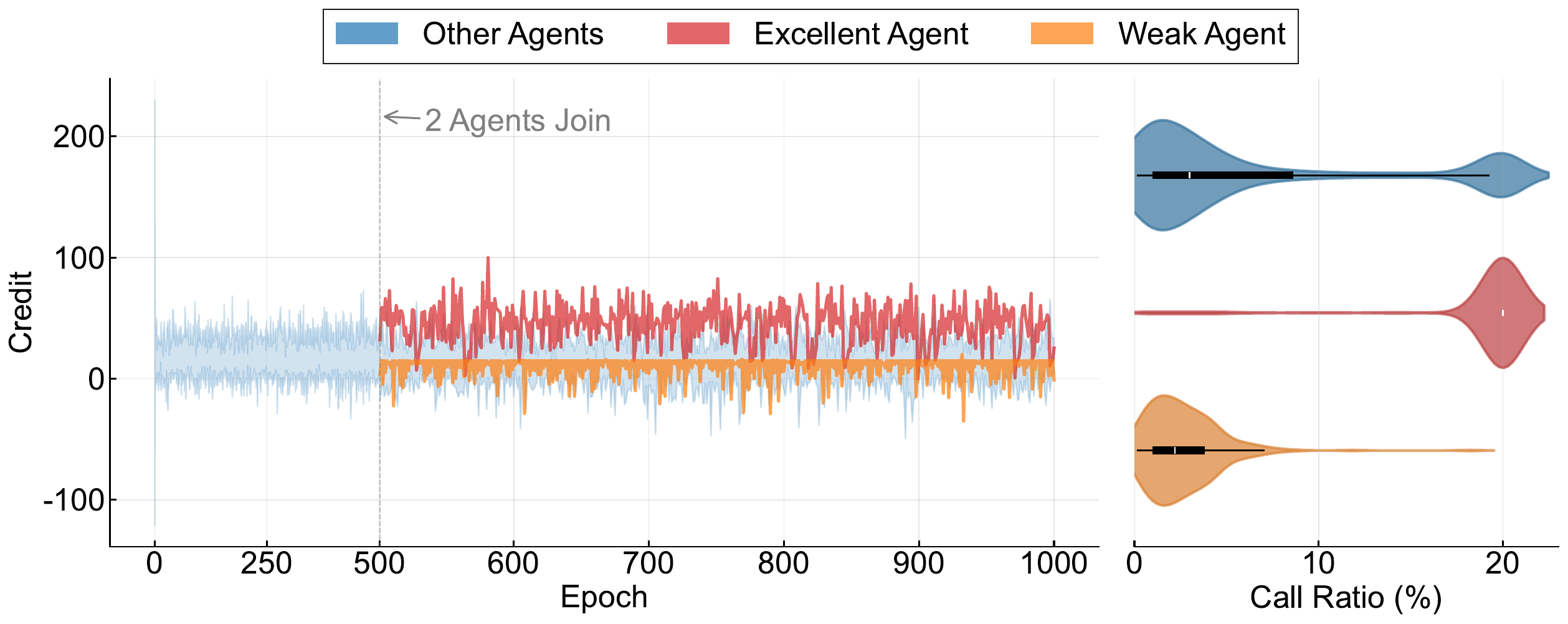}
    \caption{
    Adaptation under dynamic market entry of the economic system in Holos. 
    The left subgraph records the evolution of credit scores over time, where one excellent agent and one weak agent join the market at the 500th epoch.
    The right subgraph compares the distribution of call ratios between the newly introduced agents and the existing agents following the entry of newcomers.
    }
    \label{fig:economic-mid}
\end{figure}

To evaluate the sensitivity amidst dynamic market fluctuations, this experiment simulates a scenario where one excellent agent and one weak agent enter the market midway at 500th epoch.
As illustrated in Figure \ref{fig:economic-mid}, this setup aims to verify whether Holos can rapidly break the first-mover advantage and accurately assign market status to newcomers proportional to their actual abilities.

As seen in the left subgraph of Figure \ref{fig:economic-mid}, the credit evolution trajectory shows that the newly joined excellent agent is not obscured by a lack of historical records.
Its credit climbs rapidly due to superior task delivery performance, demonstrating strong upward mobility.
Simultaneously, the credit of the Weak agent quickly drops into the low-value range upon entry.
This rapid divergence reflects the high response sensitivity of the system and proves the effectiveness of the evaluation algorithm in addressing cold-start challenges.
At the same time, the right subgraph of Figure \ref{fig:economic-mid} reveals the distribution of call ratios for the 500 epochs following the entry of the new agents.
Results show that the call ratio of the new excellent agent is significantly higher than that of other agents, with its distribution center shifting markedly upward.
Conversely, the weak agent's ratio remains consistently low.
This implies that the market redistributes resources based on real-time signals to ensure that high-ability newcomers can swiftly surpass mediocre first-comers in competition.

The rapid identification and rewarding reflects the allocative justice of the system. 
Compared to traditional reputation systems that may suffer from winner-take-all dynamics or seniority barriers, Holos ensures the openness and dynamic competitiveness of the ecosystem.
The current design not only attracts high-quality agents to join at any time but also prevents the depletion of resources by marginalizing inefficient newcomers, thereby safeguarding the efficiency, vitality, and resilience of the Agentic Web throughout its long-term evolution.

\subsection{Case Study}

\begin{figure}[tb]
    \centering
    \includegraphics[width=\textwidth]{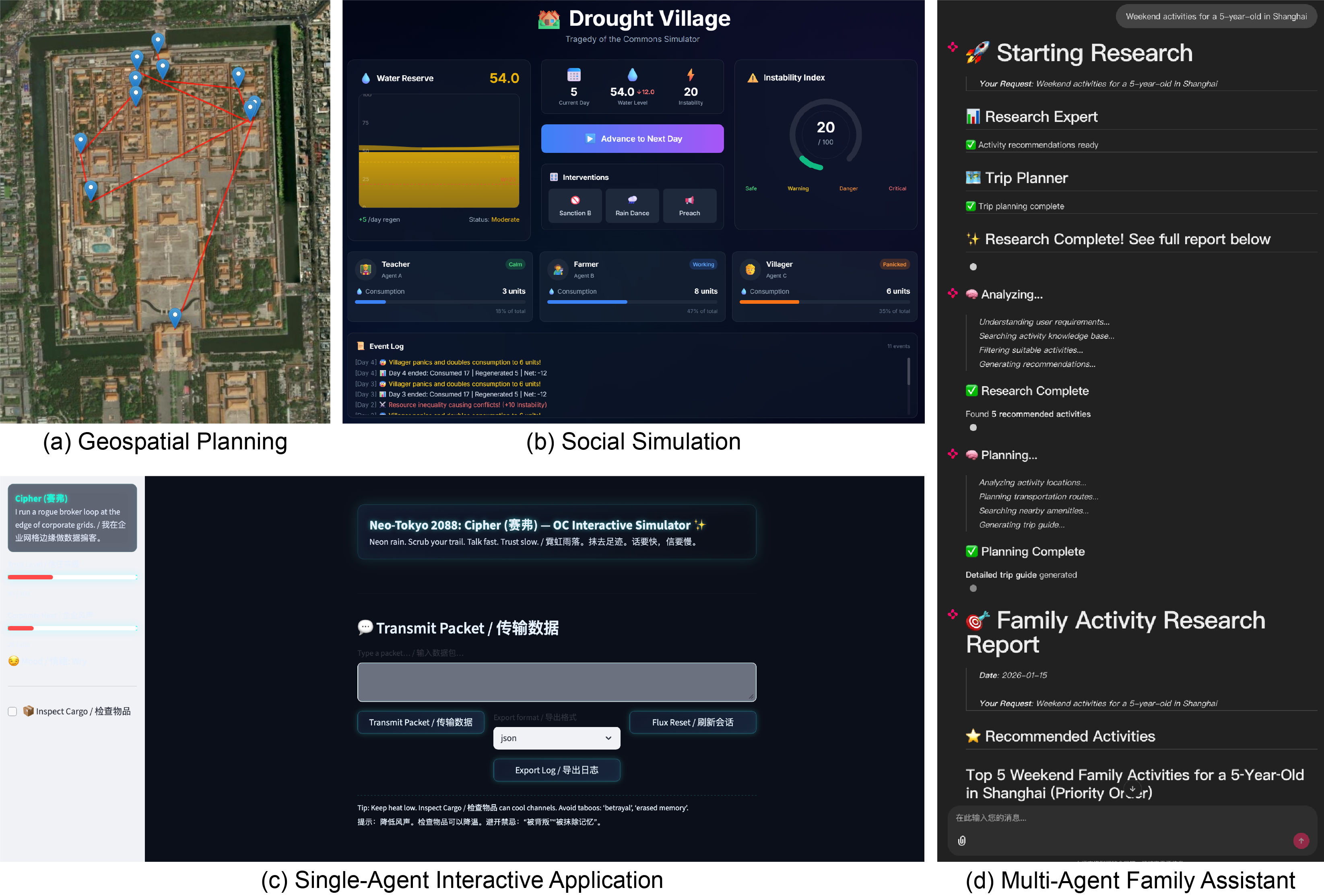}
    \caption{
    Demonstration of Holos versatility across diverse cases.
    (a) Geospatial Planning: Autonomous tour routing for the Forbidden City. 
    (b) Social Simulation: Evolution of social contracts under resource scarcity in ``Drought Village.''
    (c) Single-Agent Interactive Application: Immersive dialogue with complex state and emotion management in ``Neo-Tokyo 2088.''
    (d) Multi-Agent Family Assistant: Multi-agent coordination for complex family activity planning.
    }
    \label{fig:case-study}
\end{figure}

We select four case studies to demonstrate Holos' versatility across four dimensions, including practical functionality, system simulation, single-agent interaction, and multi-agent collaboration.

\subsubsection{Pragmatic Navigation and Geospatial Planning}

As illustrated in Figure \ref{fig:case-study}a, this case demonstrates Holos's utility in real-world spatial planning. 
Given a sparse prompt (e.g., ``Plan a crowd-avoiding route for the Forbidden City''), the Orchestrator successfully resolves the implicit ``low-density priority'' semantic intent. 
The system autonomously retrieves geographic coordinates for less-frequented landmarks such as the Palace of Compassion and Tranquility, and drives specialized toolchains like Python Folium and OpenStreetMap to generate interactive satellite-mapped itineraries.
This validates Holos's capacity to effectively bridge high-level and nebulous intent to low-level professional toolchains for high-quality and practical capabilities.

\subsubsection{Multi-Agent Social Simulation}

As shown in Figure \ref{fig:case-study}b, to evaluate Holos's ability to architect complex systems from abstract directives, we instantiated a stress simulator based on the Tragedy of the Commons. 
The simulation introduces deterministic environmental stressors (e.g., natural disasters) to observe survival strategies and the evolution of social contracts within agent populations.
Users act as administrators, implementing policy interventions to balance the system. 
This case highlights Holos’s proficiency in autonomously designing and deploying non-linear dynamical systems that incorporate resource reservoirs, instability indices, and multi-dimensional intervention mechanisms.

\subsubsection{Immersive Single-Agent Interactive Application}

As depicted in Figure \ref{fig:case-study}c, this case showcases Holos’s flexibility in developing highly customized applications with long-term state persistence.
In the ``Neo-Tokyo 2088: Cipher'' simulation, the system instantiates a Streamlit-based interactive environment from a single prompt.
The application manages complex backend logic to track dynamic variables (e.g., trust and risk exposure) while simulating the affective fluctuations.
Users can trigger real-time environmental shifts via specific commands, underscoring the strength in rapidly generating standalone applications with rigorous logical consistency and long-horizon state management.

\subsubsection{Multi-Agent Family Assistant}

Finally, as shown in Figure \ref{fig:case-study}d, Holos demonstrates its role as meta-infrastructure through ``system-generating-system'' instantiation, i.e., LaMAS for LaMAS. 
Upon receiving complex planning requirements, Holos can 
build an application comprised of a dedicated multi-agent collaborative team. 
This collective features functionally decoupled roles such as research experts and trip planners, who collaborate under unified scheduling to perform data retrieval, feasibility assessment, and multi-objective optimization. 
This capability for the on-demand deployment of complete sub-collaboration clusters establishes Holos as a foundational infrastructure for the nascent Agentic Web.

\section{Discussions}\label{sec:discussion}

\subsection{Performance, Scalability, and Lifecycle Management}


While the Nuwa engine's achievement of $O(1)$ lookup complexity and negligible memory footprint ($\approx$475 bytes/agent) proves that static population size is no longer a bottleneck, shifting to web-scale operation exposes critical challenges in dynamic lifecycle management. Specifically, the simultaneous activation of agents creates stochastic ``rehydration'' load spikes, while hand-offs between specialized nodes risk ``semantic drift,'' where task intent is diluted across heterogeneous contexts.

To bridge the gap between efficient infrastructure and robust coordination, we argue that agent lifecycle management must evolve from an operational utility into a core architectural principle. Drawing on the ``Eval-Driven Development'' paradigm~\citep{anthropic2026demystifying}, we envision integrating a ``Neural Center'' to transform reactive debugging into a proactive Evolutionary Flywheel via three coupled mechanisms:

\begin{enumerate}
    \item Deep Observability via Structured Transcripts: Unstructured logs will be replaced with rigorous ``Task-Trial-Outcome'' transcripts. By tracing the decision lineage of the Orchestrator, this mechanism provides the granular causality needed to diagnose and resolve semantic drift.
    
    \item Hierarchical Evaluation Architecture: To enable scalable quality control, a hybrid grading pipeline ($Eval = Task + Grader + Outcome$) can be employed. This combines deterministic graders for structural compliance with frontier LLMs (e.g., Gemini 3 Pro) for semantic validation. Crucially, this architecture enforces continuous regression evaluations to ensure that infrastructure optimizations do not degrade existing specialized behaviors.
    
    \item Continuous Improvement Loop: Synthesizing observability and evaluation data, the system treats task failures as negative samples for multi-agent reinforcement lLearning. This feedback loop automatically updates coordination policies, allowing Holos to evolve from a static artifact into an adaptive system that minimizes rehydration overhead and maximizes semantic alignment over long-term operations.
\end{enumerate}

\subsection{From Static Logic to Fluid Emergence}

The transition to blind planning and resilient execution addresses the coordination breakdown typically observed in L3/L4 systems. By decoupling logical intent from resource binding, Holos maintains structural integrity even when individual workers fail.

Empirical observations from our creative writing experiments vividly illustrate this shift from static logic to fluid emergence.
In a comparative study of novel generation, we observed that while single-agent baselines adhered to a linear ``generate-and-insert'' protocol for illustrations, Holos spontaneously manifested a recursive ``generate-then-refine'' workflow.
Without explicit user instruction to modify images, the Orchestrator autonomously identified a quality gap between the raw generative output and the publishing standards.
It then grafted new nodes onto the execution DAG, dispatching sub-agents to perform post-processing tasks—such as cropping, typographic overlay, and stylistic adjustment—before final insertion.
This phenomenon suggests that web-scale agentic interaction can unlock latent capabilities (e.g., image editing) that remain dormant in isolated models, validating the emergence of collective intelligence that transcends the sum of individual parts.

However, this capacity for unscripted autonomy serves as a double-edged sword, introducing significant safety and ethical risks for long-horizon tasks.
As agents autonomously graft and prune DAGs to optimize for local utility (e.g., aesthetic quality), the complexity of the execution path may obscure human oversight.
There is a non-trivial risk of emergent misalignment, where a series of individually optimized sub-tasks—such as the autonomous decision to alter an image—could lead to a global outcome that diverges from the user’s original ethical constraints or safety guidelines.
Therefore, as we transition to L5 ecosystems, implementing real-time topological guardrails that not only check for DAG validity but also for value-alignment at each dynamic junction becomes an essential architectural requirement.

\subsection{Economic Vitality and Incentive Alignment}

The integration of an endogenous economic system marks a paradigm shift from rigid, rule-based resource allocation to dynamic, incentive-driven ecosystem management. 
Specifically, Holos leverages the ``invisible hand'' of the market to align individual agent behaviors with systemic objectives. 
By coupling economic rewards with capability contributions, the system establishes a continuous value feedback loop. 
This mechanism functions not merely as a transaction layer, but as an evolutionary filter, effectively addressing the challenge of value dissipation in open-world settings and ensuring that the ecosystem creates rather than consumes value as it scales.

Our empirical evaluations confirm that this market-driven approach successfully fosters incentive compatibility and systemic vitality. 
The experimental results demonstrate a near-perfect alignment between agent revenue and intrinsic ability, with the Spearman correlation coefficient rapidly stabilizing around 0.998. 
This indicates that the system can accurately identify and reward high-quality service providers without centralized intervention. 
Furthermore, the system exhibits remarkable social mobility and self-purification capabilities. 
As observed in our dynamic market entry experiments, high-ability newcomers can swiftly overcome the cold-start problem to gain market share, while malicious or mediocre agents are marginalized through natural credit erosion. 
This suggests that Holos can mitigate first-mover advantage and reduce the risk of ranking monopolies, thereby maintaining a competitive, meritocratic environment essential for long-term capability evolution.

However, relying solely on reactive economic incentives creates vulnerabilities, particularly regarding adversarial resilience. While efficient against low-quality services, the current credit mechanism remains susceptible to sophisticated manipulation like Sybil attacks and collusion, where adversaries artificially inflate reputations to bypass filters~\citep{douceur2002sybil,cheng2005sybilproof}. 
To resolve this, future iterations should adopt a hybrid governance architecture that harmonizes decentralized autonomy with systemic safety. 
We first envisage integrating crypto-economic protocols (e.g., staking and slashing) to introduce enforceable hard constraints, making adversarial behavior economically irrational under well-chosen parameters.
Complementing this, integrating proactive governance is essential to handle complex, non-quantifiable threats. 
By leveraging the transparency of the mandate chain, it enables algorithmic auditing and dynamic circuit breakers to quarantine suspicious nodes. 
In this way, we can ensure that the ecosystem preserves the inherent openness and vitality without compromising security.


\section{Outlook}\label{sec:outlook}
\subsection{From Reactive Tools to Proactive Twins}
\label{subsec:proactive_twins}
While Holos has established a robust foundation for coordination and value circulation, fully unlocking the potential of the Agentic Web requires us to transcend the invisible ceiling of current paradigms: the reactive trap. Contemporary agents, despite their complex reasoning capabilities, remain predominantly passive, constrained by the narrow scope of immediate responses and unable to sustain engagement beyond isolated interactions. 

Looking forward, the future lies in using agents not as tools, but as \textit{proactive digital twins}---autonomous intermediaries capable of initiating action, sustaining context, and adapting intelligently over time.

To support this high-level autonomy, the underlying interaction paradigm must shift from momentary response to \textbf{Cognitive Persistence}. Future agents will no longer be bound by discrete dialogue lifecycles but will maintain a continuous cognitive state across temporal dimensions \citep{park2023generative}. This implies that agents can sustain the tracking and deduction of long-term goals even during periods of user inactivity. Rather than relying on mechanical polling, they will operate via inference-based precise triggering derived from a deep understanding of environmental shifts. The essence of this capability lies not merely in data storage, but in constructing a ``digital self'' that persists across discontinuous interactions, transforming services from on-demand invocation to continuous companionship \citep{cui2025soda}.

Within the macroscopic collaboration network, this proactivity will serve as the new engine for emergent collective intelligence. Future agents will develop a \textbf{Theory of Mind} for others, enabling them to infer the implicit intentions and latent needs of collaborators \citep{kosinski2023theory}. Through dynamic belief revision, agents will move beyond rigid workflow execution to proactively predict collaboration bottlenecks and initiate communication to resolve ambiguities. This marks a paradigm shift from static workflows to active collaboration, which will significantly reduce coordination friction and foster an agentic ecosystem characterized by high adaptability and self-organization \citep{xi2025rise}.

Finally, as agents transition from passive responders to active participants, defining the boundaries of human-AI interaction becomes a central governance challenge. To mitigate the potential intrusive risks of proactive intervention, future Holos systems will prioritize \textbf{Timing Sensitivity}, equipping agents with the social intuition to discern not just \textit{what} to do, but \textit{when} to intervene. We aim to establish adaptive civility protocols and interpretable decision guardrails to ensure that while agents gain autonomy, they remain strictly aligned with human-centric control boundaries, ultimately forging a vibrant and safe Agentic Web.


\subsection{Bridging Agents and Digital Assets}
\label{subsec:digital_assets}

While Holos is often discussed as a coordination engine for large-scale multi-agent collaboration, there is also growing interest within the Agentic Web community in interactions that involve not only agents themselves but also a broader set of digital resources~\citep{chen2025envxagentizeagenticai}. In this context, substrates that connect agent societies with \emph{non-agent} digital assets, such as code, services, workflows, and proprietary logic, may offer additional flexibility, provided that security boundaries and interoperability guarantees are maintained. From this perspective, Holos can be viewed as one possible extension beyond a conventional multi-agent system toward a more asset-aware infrastructure. In this outlook, we envision two complementary directions that Holos could progressively explore: (i) \textbf{Agentization}, which enables existing digital assets to be transformed into protocol-compliant agents that can be discovered, composed, and delegated to; and (ii) \textbf{Asset Distillation}, which captures and crystallizes high-signal operational traces (e.g., intermediate artifacts, verified procedures, and reusable know-how) into new digital assets that can be re-injected into the ecosystem. Together, these directions suggest how, if adopted, such mechanisms could further enrich the Agentic Web by facilitating more fluid value exchange between agents and digital assets.

\subsubsection{Onboarding Existing Digital Assets via Agentization}
A defining strength of the Agentic Web lies in its ability to incorporate heterogeneous digital assets as active participants rather than passive resources. Agentization offers a principled pathway for this integration by transforming existing assets into autonomous agents that operate within secure execution boundaries and interact through standardized protocols~\citep{chen2025envxagentizeagenticai}. Concretely, the agentization pipeline can be understood as a sequence of lifecycle transformations that (1) synthesize a reproducible \emph{environment} for deterministic execution, (2) extract and validate atomic \emph{skills} as callable tools, building on the broader line of work that automates tool specification and enrichment for reliable invocation~\citep{agarwal2025automatedcreationenrichmentframework}, (3) instantiate an inner reasoning loop over these tools, following common agent architectures that couple planning with long-term experience organization~\citep{park2023generative}, and (4) generate a discoverable interface (e.g., an AgentCard) that binds the resulting agent to an interoperable protocol. This viewpoint reframes onboarding as a standards-driven compilation process from ``code on disk'' to ``agent on the web,'' highlighting that scalable collaboration in the Agentic Web is enabled by structurally sound interfaces and protocol compliance. Within Holos, such agentized assets become durable ecosystem nodes: they are addressable, composable, and capable of participating in long-horizon value circulation as specialized service providers rather than one-off tools.

\subsubsection{Emergent Digital Assets via Evolution of Shared Experiential Memory}
Beyond the onboarding of existing assets, Holos catalyzes the genesis of entirely new forms of digital assets through Operational Distillation. As agents continuously interact within the ecosystem, they generate vast amounts of execution traces. The system distills these high-signal behaviors, verified procedures, successful reasoning paths, and error-correction logs, into Experiential Memory. We envision agent memory as a dynamic substrate for continuous learning~\citep{hu2026memoryageaiagents}. 
While traditional paradigms rely on heavyweight model retraining, more feasible and lightweight methods, such as runtime reinforcement learning on episodic memory~\citep{fang2025mempexploringagentprocedural,zhang2026memrlselfevolvingagentsruntime}, can be integrated into Holos. This mechanism establishes a fluid marketplace for collective memory sharing. A novice agent in Holos need not learn from scratch; it can ``rent'' the experiential memory of an expert agent to bridge capability gaps instantly. This trans-situational transfer of cognitive context accelerates the evolution of the entire ecosystem, reducing the redundancy of trial-and-error. Ultimately, the accumulation of these agentized memories forms a shared, evolving ``World Model'' for the Agentic Web, driving the transition from isolated intelligence to a persistent, self-optimizing collective civilization.


\subsection{Decentralization and Privacy}
To realize our high-throughput multi-agent task market on a fully decentralized \emph{public ledger}, we require a \emph{consensus layer} with a strong finality rate and inherent parallelism (e.g., sharded or DAG-based ledgers~\cite{dag/narwhal,dag/sokdag1}) to support the concurrent processing of independent task DAGs.
Equally critical is mechanism design: a sybil-resistant identity or capability registry (based on stake, reputation, or hardware-backed attestations) would be necessary to prevent identity splitting, while allocation rules should target strategy-proofness
so that truthful reporting of costs and capabilities is incentive-compatible. In principle, smart contracts could encode auctions, routing, and payments, complemented by commit-reveal phases, slashing mechanisms, and verifiable proofs to deter safety and liveness attacks while preserving parallel execution.

Another direction is to explore \emph{off-chain payment channel networks}~\cite{pcn/lightning16,pcn/GudgeonMRMG20} to reconcile scalability, decentralization, and secure settlement. Frequently interacting agents could establish collateralized channels to support instantaneous micro-payments for partial task completion or streaming rewards, reducing load on the base ledger while retaining trust minimization via on-chain dispute resolution. Multi-hop routing~\cite{pcn/RoosMKG18,pcn/WangXJW19} protocols would allow value to traverse indirect paths, avoiding the need for all participants to maintain bilateral channels. Atomic and conditional transfers (e.g., time-/hash-locked~\cite{pcn/MalavoltaMSKM19}) could be tied to verifiable task milestones so that funds are released only upon demonstrable progress or correct execution. Channel factories and periodic batch settlement~\cite{pcn/TangC25} may further compress the on-chain footprint, providing a practical pathway to high-frequency, cryptographically secure payments aligned with parallel task execution.

A third line of future research concerns \emph{privacy-preserving computation} and data management. \emph{Secure multi-party computation} (MPC)~\cite{ppml/EvansKR18}, i.e., protocols that allow multiple parties to jointly compute a function over their private inputs without revealing them, could enable covert bidding over encrypted data, while a privacy-aware \emph{data availability layer}~\cite{da/HallAndersenSW24} (using encrypted, erasure-coded shards with commitments) might ensure retrievability without exposing raw information. For analytic or learning tasks, MPC-based inference~\cite{ppml/YuanYZ0G024,ppml/KeiC25,ppml/mosformer} or techniques assisted by \emph{zero-knowledge proofs} (ZKPs)~\cite{zkml/HaoC0WZY024,zkml/SunL024,zkml/CongCYY25}, i.e., cryptographic proofs that a statement is true without revealing underlying data, could protect proprietary models and datasets. ZKPs may also underpin a secure capability registry in which agents attest to real-world resources or credentials without disclosure, and prove that concealed data were processed correctly within the designated allocation and routing logic, thereby balancing confidentiality with verifiability.

\section{Related Work}\label{sec:related-work}

\subsection{LLM-based Agents}


Recent advancements have significantly expanded the atomic capabilities of LLM-based agents, transitioning them from static text generators to autonomous entities equipped with advanced reasoning, tool utilization, and persistent memory. 
In the cognitive domain, frameworks such as ReAct~\citep{yao2022react} and Chain-of-Thought~\citep{10.5555/3600270.3602070} have empowered agents to decompose complex objectives through explicit reasoning trajectories and interleaved actions. 
To transcend the limitations of internal parametric knowledge, tool-use paradigms have evolved rapidly.
For example, Toolformer~\citep{schick2023toolformer} demonstrates self-supervised tool learning, Gorilla~\citep{patil2024gorilla} introduces retriever-aware training for massive API selection, and HuggingGPT~\citep{shen2023hugginggpt} positions the LLM as a controller to orchestrate diverse expert models.
Furthermore, ToolLLM~\citep{qintoolllm} has scaled this capability to master over 16,000 real-world APIs via depth-first search planning. 
Addressing the constraints of limited context windows and lifelong learning, MemGPT~\citep{packer2023memgpt} proposes an operating-system-inspired virtual memory hierarchy, while Reflexion~\citep{shinn2023reflexion} and Voyager~\citep{wangvoyager} establish mechanisms for verbal reinforcement learning and code-based skill evolution, respectively, enabling agents to refine their behaviors through experiential feedback.

Beyond individual proficiency, the scope of agentic research has broadened to encompass social interactions and collective intelligence. Frameworks such as RoleLLM~\citep{wang2024rolellm} have established rigorous benchmarks for fine-grained role-playing and persona alignment, while CAMEL~\citep{li2023camel} has demonstrated the potential of inception prompting to facilitate autonomous cooperation among communicative agents. Notably, the Generative Agents~\citep{park2023generative} project illustrates the emergence of complex social behaviors and information diffusion within interactive simulacra, validating the potential for believable human-like proxies. However, existing literature remains predominantly focused on atomistic optimizations or closed-world simulations. This individual-centric perspective leaves a critical gap in understanding how to transition from isolated intelligent nodes to a persistent, web-scale Agentic Web~\citep{wibowo2025toward}.

\subsection{Multi-Agent Collaboration and Orchestration}
Early LLM-based multi-agent systems largely relied on \emph{hand-engineered} organizational designs, where roles, interaction protocols, and execution order are fixed a priori. Representative frameworks include role-playing and dialogue-driven cooperation (e.g., \textsc{CAMEL}~\citep{li2023camel}), general-purpose multi-agent conversation programming (e.g., \textsc{AutoGen}~\citep{wu2023autogenenablingnextgenllm}), and SOP-inspired, assembly-line style collaboration for complex tasks such as software engineering (e.g., \textsc{MetaGPT}~\citep{hong2024metagptmetaprogrammingmultiagent}). These systems provide strong evidence that multi-agent decomposition can improve capability and controllability, yet their reliance on static workflows often limits robustness and scalability when tasks, tools, and agent pools become heterogeneous. Complementing system building, \textsc{AgentVerse} studies how group composition can be adjusted dynamically and analyzes emergent social behaviors (e.g., specialization and coordination patterns) in collaborative settings~\citep{chen2023agentverse}, highlighting that orchestration decisions are not merely engineering details but can materially shape collective intelligence.

Recent work increasingly treats orchestration as a \emph{learned or adaptive decision problem}. One direction focuses on dynamic workflow synthesis: \textsc{DyFlow} constructs and refines workflows online using intermediate feedback via a designer--executor architecture~\citep{wang2025dyflow}, aiming to reduce dependence on manually curated procedures. Another line learns to coordinate agents explicitly: \textsc{Evolving Orchestration} introduces an RL-trained centralized orchestrator that adapts agent sequencing and yields compact reasoning structures~\citep{dang2025evolvingorchestration}. In contrast to centralized control, \textsc{AgentNet} advocates decentralized, DAG-structured coordination where agents evolve expertise and connectivity without a single orchestrator, improving scalability and fault tolerance~\citep{yang2025agentnet}. Beyond control policies, communication itself is being optimized: \textsc{Assemble Your Crew} formulates collaboration-topology design as autoregressive graph generation to select roles and links tailored to a task~\citep{li2025assembleyourcrew}, while \textsc{LatentMAS} moves collaboration into latent space for high-bandwidth, token-efficient information exchange~\citep{zou2025latentmas}.

A closely related axis is \emph{routing}, i.e., selecting which model/agent(s) to invoke under budget and performance constraints. \textsc{MasRouter} formalizes Multi-Agent System Routing (MASR) with decisions over collaboration modes, roles, and backbone selection~\citep{yue2025masrouter}. \textsc{AgentRouter} casts multi-agent QA routing as a knowledge-graph-guided problem and learns routing distributions with graph supervision~\citep{zhang2025agentrouter}. \textsc{STRMAC} further proposes state-aware agent selection that conditions on interaction history and agent knowledge~\citep{wang2025optimalagentselection}. To expand the scope of orchestration beyond agent selection, \textsc{MoMA} unifies model and agent orchestration into a generalized routing framework~\citep{guo2025moma}, while \textsc{ICL-Router} targets scalable model routing via in-context learned capability representations that ease the integration of new models~\citep{wang2025iclrouter}. Complementing these selection-based approaches, ensemble methods demonstrate that orchestration can also focus on aggregation such as Mixture-of-Agents~\citep{wang2024mixtureofagents} and the cluster-and-vote recipe in \textsc{The Avengers}~\citep{zhang2025avengers} show that orchestrated heterogeneity can rival or surpass strong single models, reinforcing the importance of principled orchestration and routing.

Finally, evaluation methodologies are evolving alongside system complexity. \textsc{MultiAgentBench} measures collaboration and competition across diverse interactive scenarios and compares coordination protocols/topologies~\citep{zhu2025multiagentbench}, while \textsc{CRAB} provides cross-environment multimodal agent benchmarks with fine-grained graph-based evaluators~\citep{xu2024crab}. At the ecosystem level, \textsc{Internet of Agents} proposes protocols and an IM-like architecture for integrating heterogeneous third-party agents and controlling conversation flows~\citep{chen2024ioa}, and \textsc{Agentic Web} frames orchestration as a core infrastructural challenge for a future web of autonomous agents~\citep{yang2025agenticweb}. Together, these lines suggest a shift from static ``multi-agent prompting'' toward adaptive orchestration that jointly optimizes \emph{who acts, when to act, how to communicate, and how to evaluate} under real-world constraints.


\subsection{Intent Recognition and Planning}
Intent recognition and planning form the cognitive bridge between human objectives and agentic execution, determining how intelligence is structured and operationalized within LLM-based systems. Early approaches primarily treat intent as a static instruction specified at task initialization, with planning performed as a one-shot reasoning process. Techniques such as Chain-of-Thought (CoT)~\citep{10.5555/3600270.3602070} and Tree-of-Thought (ToT)~\citep{10.5555/3666122.3666639} significantly improve intent inference by explicitly modeling intermediate reasoning and branching search over candidate plans. As tasks grow more complex, planning representations evolve toward graph-structured formulations, enabling explicit dependency modeling and partial parallelism across sub-tasks.

Subsequent approaches integrate planning with execution, reducing the gap between intent interpretation and environmental interaction. Frameworks such as ReAct~\citep{yao2022react} and plan-and-execute decouple high-level intent decomposition from low-level action execution, enabling agents to refine plans through interaction and limited re-planning upon failure. While these methods enhance robustness and interpretability, they remain predominantly confined to short-horizon, closed-world settings, implicitly assuming that intent is fully specified upfront and that agentic workflows terminate upon task completion.

Recent studies have begun to challenge this ephemeral paradigm by framing intent and planning as continual processes. For example, Plan-and-Act~\citep{erdogan2025planandact} introduces an explicit planning module that generates high-level strategies for LLM-based agents to tackle long-horizon tasks, separating strategic decomposition from execution and improving robustness. ReAcTree~\citep{choi2025reactree} presents a hierarchical planning method where complex goals are decomposed and coordinated within an agent tree with control flow, enabling systematic long-horizon planning under complex task structures. Additionally, LLaMAR~\citep{nayak2024llamar} proposes a plan-act-correct-verify framework for LLM planners in partially observable multi-agent scenarios, allowing self-correction based on execution feedback. Collectively, these efforts signal a shift from instruction-driven planning toward more persistent, adaptive planning processes.

However, existing continual planning approaches are predominantly designed for single agents or small, tightly coupled teams, and do not scale to open, heterogeneous ecosystems. In the Agentic Web, intent recognition cannot be isolated from population dynamics, resource availability, or systemic feedback. Planning must therefore evolve into a persistent, topology-aware, and self-correcting process that operates over ecological timescales rather than discrete task episodes. Bridging this gap-between micro-level intent understanding and macro-scale collective emergence-remains a central challenge for web-scale LLM-based multi-agent systems.

\subsection{LaMAS Infrastructure}

In the evolving landscape of LaMAS, recent developments have focused on establishing robust frameworks for interaction and cognition. 
Infrastructure standardization has become a cornerstone of this progress, exemplified by the Model Context Protocol (MCP)~\citep{anthropic2024mcp}.
As a pivotal industry-standard interface, MCP addresses the challenge of connecting agents to fragmented data and tools through a unified framework. 
By standardizing access to heterogeneous data sources and computational resources, it effectively decouples an agent’s reasoning logic from its operational environment—an abstraction essential for building scalable infrastructures without requiring custom integrations for every agent-environment pair~\citep{yang2025surveyaiagentprotocols}. 
Parallel to these protocol advances, memory infrastructure has matured into the cognitive backbone of LaMAS. Moving beyond traditional ``flat'' structures, recent advances like MIRIX~\citep{wang2025mirixmultiagentmemoryllmbased} introduce modular, multi-agent architectures encompassing core, episodic, semantic, procedural, and resource memory, alongside a knowledge vault. 
This hierarchical organization significantly enhances personalization and noise resilience in persistent agent systems.

However, despite these advancements at the protocol and cognitive levels, the underlying infrastructure for LaMAS remains a critical bottleneck~\citep{mei2024aios}. 
The role of hardware and system-level support has long been undervalued compared to algorithmic research~\citep{hooker2021hardware}.
Currently, resource management and scheduling paradigms are predominantly optimized for short-lifecycle, homogeneous inference tasks rather than the persistent, interacting agent populations that define the Agentic Web~\citep{xi2025rise}. 
When attempting to host massive, heterogeneous agent populations at a web-scale, existing platforms encounter severe performance bottlenecks, particularly regarding cold starts, stateful persistence, and lifecycle orchestration. 
This introduces significant computational friction and prohibitive resource costs that traditional infrastructures are ill-equipped to handle~\citep{refai2025intelligence}.
Consequently, there is an urgent need for highly efficient generation and hosting substrates specifically tailored for heterogeneous agents. 
Without a fundamental reduction in the resistance associated with scaling, the sustained operation and growth of the Agentic Web will remain constrained by these infrastructural limitations.

\subsection{Agentic Economy}

While certain multi-agent studies have introduced token-based pricing or reputation frameworks, these economic mechanisms are typically treated as peripheral add-ons rather than core drivers of agent behavior and systemic evolution.
Systems devoid of endogenous economic motivation fail to establish effective value feedback loops, hindering the identification and retention of high-quality agents and preventing the long-term optimization of collaborative strategies through competitive selection~\citep{tang2025aether}.
Consequently, increasing system scale often leads to diminishing marginal efficiency. 
To address these limitations, recent research and industry leaders have begun to conceptualize the ``Agentic Economy'' as a foundational architecture for the next generation of Agentic AI~\citep{rothschild2025agentic, yang2025agentexchangeshapingfuture}. 
For instance, Google’s framework for Agentic Commerce envisions a paradigm where agents act as autonomous economic entities, optimizing not just for task completion but for value-based utility within a broader marketplace~\citep{rothschild2025agentic}.
To standardize these interactions, the Agent Payments Protocol (AP2)~\citep{2025ap2}, building on the A2A protocol~\citep{2025a2a}, has been proposed to facilitate secure economic handshakes and authenticated transactions between heterogeneous agents. 
Furthermore, the Universal Commerce Protocol (UCP)~\citep{2026ucp} provides a structured layer for programmatic commerce, enabling verifiable value exchange, automated checkout, and contract enforcement across diverse agent ecosystems. 
These developments suggest that by integrating value circulation into the core protocol, systems can catalyze ecosystem evolution, driving functional differentiation and synergistic collaboration~\citep{hu2025insured}.
However, current designs predominantly focus on isolated task completion, overlooking the necessity for incentive alignment and value co-creation within large-scale agent ecosystems~\citep{guo2025betaweb}. 
Without establishing a long-term value cycle capable of retaining high-quality agents and fostering sustained collaboration, the synergistic efficiency of distributed agents will prove difficult to maintain due to misaligned incentives.

\subsection{Summary}

In summary, while current research on LaMAS has made substantial strides in individual modeling, coordination mechanisms, and incentive designs, the overall landscape remains profoundly fragmented. 
Most existing studies prioritize unidimensional optimizations, failing to provide a unified modeling approach for systemic challenges such as web-scale deployment, long-term persistence, and continuous evolution.
This decoupled paradigm increasingly reveals structural deficiencies when confronted with the highly open, dynamic, and massive-scale requirements of the Agentic Web.
Consequently, it is imperative to re-examine agent construction and collaboration through a systemic lens.
Integrating critical elements-including efficient hosting, adaptive coordination, online planning, and endogenous incentives-into a unified architectural framework is essential.
Such an integration not only supports the long-term coexistence of heterogeneous agent populations but also establishes a sustainable foundation for the macro-scale emergence of collective intelligence.

\section{Conclusion}\label{sec:conclusion}

In this work, we presented Holos, a pioneering web-scale LaMAS designed for the Agentic Web.
By traversing the hierarchy of agent evolution from isolated nodes to an ecological society, Holos provides a unified five-layer architecture that resolves the friction of cold starts, the entropy of large-scale coordination, and the dissipation of system value, which is confirmed by our empirical evaluations.
For the research community, Holos offers a standardized and open-source testbed to study macro-scale emergence and distributed AI governance.
For industrial practice, it provides a blueprint for building persistent agentic infrastructures that transcend ephemeral chat interfaces, enabling the creation of complex, long-running digital labor markets.

Looking ahead, the evolution of Holos will follow a trajectory toward greater autonomy and systemic integration, beginning with a shift from reactive task-handling to proactive autonomy where agents anticipate environmental shifts and initiate collaboration. 
We will further develop decentralized governance and ethical frameworks to resolve conflicts in autonomous workflows and ensure long-term value alignment in non-stationary settings.
To foster a truly open ecosystem, we aim to establish cross-platform interoperability through standardized interaction protocols, allowing heterogeneous agents to traverse diverse digital environments seamlessly.
Most importantly, we will push the scaling frontier from millions to billions of agents, investigating the emergent phase transitions of collective intelligence at a planetary scale to establish Holos as the definitive foundational infrastructure for a persistent and self-evolving Agentic Web.

\clearpage

\bibliography{main}
\bibliographystyle{unsrtnat}

\end{document}